\documentclass[10pt,twocolumn,letterpaper]{article}

\usepackage{cvpr}              

\usepackage[dvipsnames,table]{xcolor}

\usepackage{multirow}
\usepackage{array}
\usepackage{algorithm}
\usepackage{algpseudocode}
\usepackage{standalone}   
\usepackage{tikz}
\usetikzlibrary{arrows.meta,backgrounds,calc,fit,positioning}

\definecolor{cvprblue}{rgb}{0.21,0.49,0.74}
\usepackage[pagebackref,breaklinks,colorlinks,citecolor=cvprblue]{hyperref}

\def\CurrentTheme{tech2}
\expandafter\ifx\csname CurrentTheme\endcsname\relax\def\CurrentTheme{ppt}\fi

\def\ThemePPT{ppt}
\def\ThemeTech{tech}

\ifx\CurrentTheme\ThemePPT
  \definecolor{tensor}{HTML}{595959}    
  \definecolor{trainable}{HTML}{F4B183} 
  \definecolor{operation}{HTML}{C55A11} 
  \definecolor{groupbg}{HTML}{F1F4F9}   
\else\ifx\CurrentTheme\ThemeTech
  \definecolor{tensor}{HTML}{33292F}    
  \definecolor{trainable}{HTML}{D9531E} 
  \definecolor{operation}{HTML}{32589C} 
  \definecolor{groupbg}{HTML}{F2F5F7}   
\else
  \definecolor{tensor}{HTML}{172A3A}    
  \definecolor{trainable}{HTML}{D14F1D} 
  \definecolor{operation}{HTML}{4E54C8} 
  \definecolor{groupbg}{HTML}{F3F4F8}   
\fi\fi

\def\paperID{37} 
\def\confName{3DV\xspace}
\def\confYear{2027\xspace}

\title{ARG-Avatar: Animation-Rendering Guidance for \\ Efficient Feedforward Avatar Reconstruction}

\author{
Jianjin Xu \qquad Soham Dasgupta \qquad Fernando De la Torre\\
Carnegie Mellon University, Robotics Institute\\
{\tt\small \{jianjinx,sdasgupt,ftorre\}@andrew.cmu.edu}
}

\newcommand{\xv}[0]{\mathbf{x}}

\newcommand{\yv}[0]{\mathbf{y}}
\newcommand{\zv}[0]{\mathbf{z}}
\newcommand{\muv}[0]{\boldsymbol{\mu}}
\newcommand{\Pv}[0]{\mathbf{P}}
\newcommand{\pv}[0]{\mathbf{p}}
\newcommand{\Qv}[0]{\mathbf{Q}}
\newcommand{\Kv}[0]{\mathbf{K}}

\newcommand{\Vv}[0]{\mathbf{V}}
\newcommand{\cv}[0]{\mathbf{c}}
\newcommand{\qv}[0]{\mathbf{q}}
\newcommand{\Dv}[0]{\mathbf{D}}

\newcommand{\sv}[0]{\mathbf{s}}
\newcommand{\ov}[0]{\mathbf{o}}
\newcommand{\Rv}[0]{\mathbf{R}}

\newcommand{\shv}[0]{\mathbf{SH}}

\newcommand{\nv}[0]{\mathbf{n}}
\newcommand{\fv}[0]{\mathbf{f}}
\newcommand{\Cm}[0]{\mathcal{C}}
\newcommand{\Gm}[0]{\mathcal{G}}
\newcommand{\Iv}[0]{\mathbf{I}}
\newcommand{\dv}[0]{\mathbf{d}}
\newcommand{\thetav}[0]{\boldsymbol{\theta}}

\newlength\savedwidth
\newcommand\whline[1]{\noalign{\global\savedwidth\arrayrulewidth
                               \global\arrayrulewidth #1} %
                      \hline
                      \noalign{\global\arrayrulewidth\savedwidth}}

\begin{document}
\twocolumn[{%
\maketitle
\vspace{-0.23in}
\begin{center}
\begin{tikzpicture}[inner sep=0pt,outer sep=0pt]
\node[anchor=south west,inner sep=0pt] (teaserimg) at (0,0) {\includegraphics[width=0.95\textwidth]{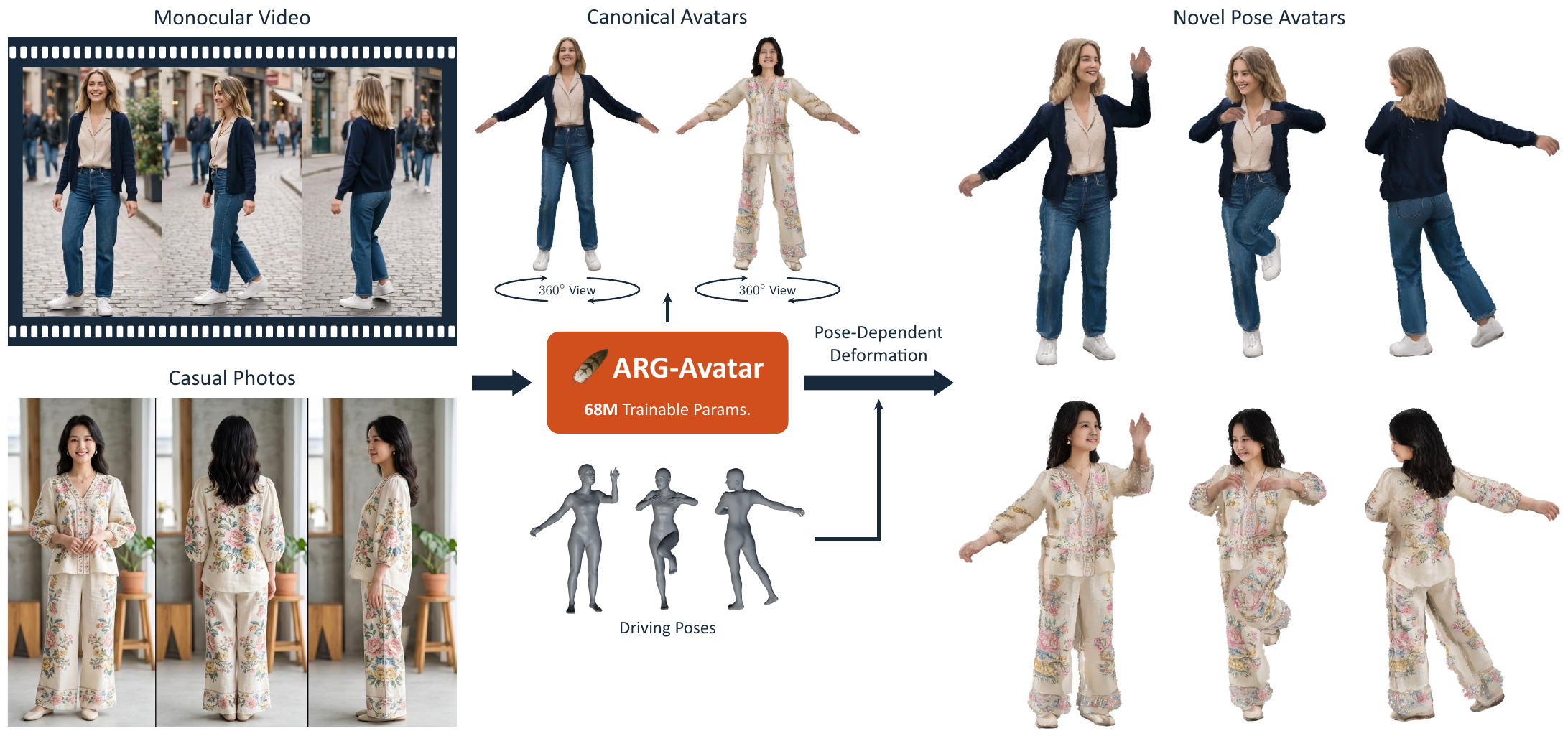}};
\end{tikzpicture}
\begingroup
\captionsetup{type=figure,aboveskip=4pt,belowskip=0pt}
\vspace{-0.05in}
\captionof{figure}{
Overview of ARG-Avatar.
Given casual photographs or frames from a monocular video, ARG-Avatar reconstructs a canonical avatar and produces avatars with pose-dependent deformations for arbitrary driving poses. ARG-Avatar has only 68M trainable parameters.}
\label{fig:teaser}
\endgroup
\end{center}
}]

\begin{abstract}
\vspace{-0.2in}
   Reconstructing animatable 3D human avatars from images is important for applications such as gaming and augmented reality.
   Feedforward methods make avatar reconstruction accessible by reducing inference time to a few seconds, but recent approaches rely heavily on scaling model capacity, training data, and computation. We ask whether such scaling is necessary.
    To this end, we introduce ARG-Avatar, an efficient feedforward framework with only 68M trainable parameters that achieves SOTA reconstruction quality.
    Our key insight is that 3D animation and rendering already provide strong geometric correspondences that existing methods largely attempt to learn from data. In light of this, we design the Foreground Avatar Coordinate RoPE attention (FACRoPE) and Intermediate Token Rendering (ITR) pipeline to guide avatar tokens to attend to their corresponding patches. Experiments across multiple datasets show that ARG-Avatar consistently outperforms prior methods when using two or more input images. Compared with LHM++, ARG-Avatar uses approximately 9$\times$ fewer trainable parameters, 4.8$\times$ fewer training sequences, and 14$\times$ less training compute, while running 6$\times$ faster and achieving a 2.7 dB PSNR improvement on the out-of-distribution NeuMan benchmark.

\end{abstract}

\section{Introduction}
\label{sec:intro}

3D avatars serve as our embodied representations in virtual environments, enabling users to appear, move, and interact in 3D while preserving their identity and motion. They are already widely used as controllable characters in games, virtual actors in film production, and digital humans for livestreaming. Their adoption could expand substantially if non-expert users could create high-quality, animatable avatars at low cost through simple capture procedures.
Ultimately, such technology could enable virtual spaces in which anyone can participate using a personalized 3D reconstruction of themselves.

Animatable avatars can be obtained through either generation or reconstruction.
Generative methods learn a distribution of human avatars conditioned on inputs such as a single image or a text description~\cite{chan2021efficient, chan2023generative, zhang2024rodinhd}.
In contrast, reconstruction methods estimate an avatar that explains a set of observed images or video frames~\cite{jiang2022instantavatar, hu2023gauhuman, wen2024gomavatar}.
This work focuses on reconstruction, with the goal of making high-quality avatar creation accessible with casual captures and modest computational resources.

Existing avatar reconstruction methods broadly fall into two categories: optimization-based and feedforward-based.
Optimization-based approaches fit a 3D representation separately to each subject using input images. Recent methods~\cite{moon2024expressive, li2024gaussianbody, wen2024gomavatar} can reconstruct avatars from monocular videos captured with ordinary smartphones, but their per-subject optimization may still require tens of minutes or even hours.
Feedforward methods~\cite{zhuang2024idol, qiu2025lhm, qiu2025lhmb}, by contrast, predict an avatar in a single network forward pass from one or a few images, reducing reconstruction time to only a few seconds.
These methods typically predict an avatar anchored to a parametric human template, allowing the resulting representation to be animated to target poses using Linear Blend Skinning (LBS). Despite their efficient inference, recent feedforward approaches have followed a scaling-first training paradigm, relying on increasingly large transformer architectures and increasingly large proprietary datasets.
For example, IDOL~\cite{zhuang2024idol} is trained on 102K sequences, while the LHM series~\cite{qiu2025lhm, qiu2025lhmb} uses approximately 300K internal videos.
This raises a fundamental question: {\bf is large-scale model and data growth necessary for high-quality feedforward avatar reconstruction?}

We challenge this scaling-first philosophy.
First, increasing model capacity cannot overcome the representational constraints imposed by a template-anchored avatar.
Second, existing architectures do not fully exploit the geometric structure already provided by the template.
Most methods rely on stacks of generic attention layers, forcing the network to learn the correspondence between avatar elements and image regions almost entirely from data. However, because each avatar element is attached to a known template and transformed by LBS, much of this correspondence can be computed geometrically rather than learned from scratch. Ignoring this prior leads to an unnecessarily data- and compute-intensive learning problem.

Motivated by this observation, we introduce \textbf{Animation and Rendering Guidance for Avatar Reconstruction}, or \textbf{ARG-Avatar}, an efficient feedforward framework that explicitly incorporates template, animation, and rendering geometry into the reconstruction network.
An overview of ARG-Avatar is shown in~\cref{fig:teaser}.
The central idea is to guide each avatar token toward the image region in which the corresponding 3D element would appear after being animated and rendered under the observed pose.
ARG-Avatar realizes this idea through two main components: Foreground Avatar Coordinate RoPE (FACRoPE) and Intermediate Token Rendering (ITR).
FACRoPE augments cross-attention with Foreground Avatar Coordinates, which encode the projected location and visibility of each avatar token after animation and rendering. ITR decodes intermediate network tokens into a coarse avatar, providing a better geometric base for animation than the template. Together, these components establish a rendering-in-the-loop mechanism that improves token-to-image correspondence without relying on large-scale model and data.

\begin{figure}[t]
    \centering
    \includegraphics[width=0.95\linewidth]{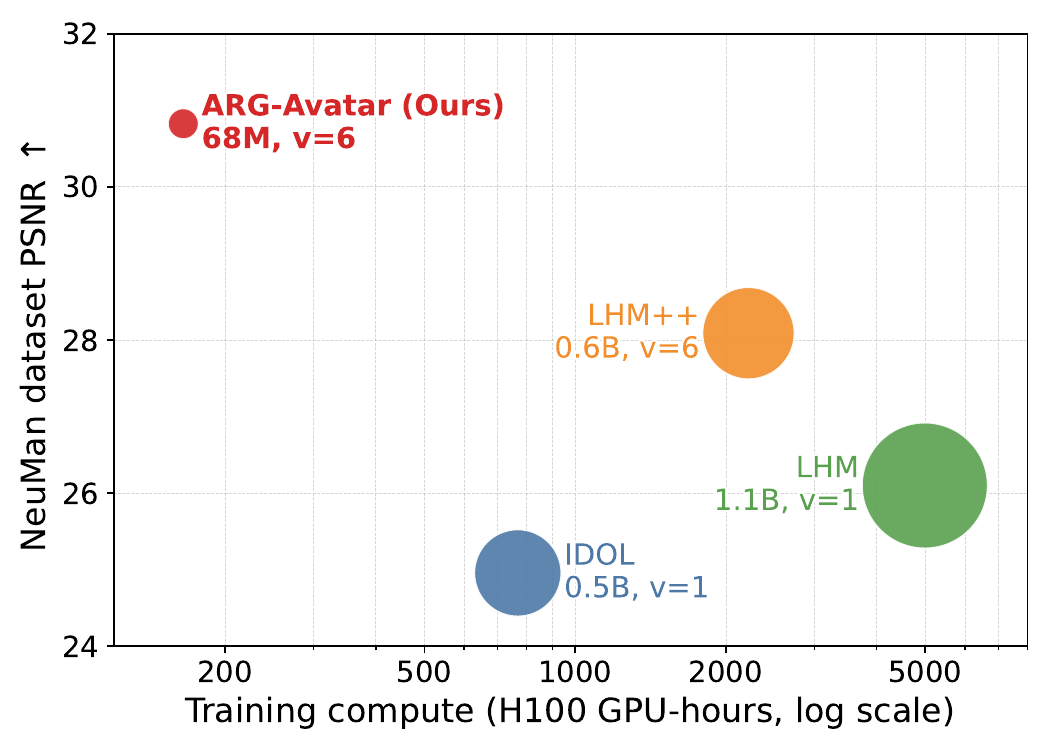}
    \vspace{-0.15in}
    \caption{
    Out-of-distribution NeuMan PSNR vs. training compute. Circle area is proportional to trainable parameters.
    }
    \vspace{-0.15in}
    \label{fig:efficiency_bubble}
\end{figure}

We train ARG-Avatar exclusively on publicly available datasets and evaluate it both on held-out validation subjects and on the out-of-distribution NeuMan benchmark~\cite{jiang2022neuman}, which is not used during training.
We compare against SIGMAN~\cite{yang2025sigman}, IDOL~\cite{zhuang2024idol}, LHM~\cite{qiu2025lhm}, and LHM++~\cite{qiu2025lhmb}.
With two or more input views, ARG-Avatar outperforms all compared methods in PSNR across the reported datasets.
Compared with LHM++, which uses 609M trainable parameters and is trained on 300K private commercial videos, ARG-Avatar uses only 68M trainable parameters and 64K public training sequences.
Nevertheless, it achieves a 2.7 dB PSNR improvement on NeuMan while running $\sim 6\times$ faster at inference time.
Figure~\ref{fig:efficiency_bubble} summarizes this trade-off.

In summary, our contributions are:
\begin{itemize}
    \item We introduce \textbf{FACRoPE}, a geometry-aware attention mechanism that uses an animatable template, LBS, and rendering to improve correspondence between avatar tokens and image features.
    \item We propose \textbf{ITR}, a rendering-in-the-loop pipeline that decodes intermediate 3D avatar predictions and uses their projected geometry to guide subsequent network layers.
    \item We develop an efficient feedforward avatar reconstruction framework that achieves state-of-the-art quality and efficiency, reaching 30.82 dB PSNR on NeuMan with only 68M trainable parameters and substantially less training data and computation than prior methods.
\end{itemize}

\section{Related Work}
\label{sec:related}

\paragraph{Optimization-based avatar fitting.}
Animatable human reconstruction is usually organized around articulated body templates such as SMPL-X~\cite{loper2023smpl,pavlakos2019expressive,ferguson2025mhr,park2025atlas,guo2024makeitanimatable}.  Per-subject methods optimize avatar attributes so that the rendered subject explains the captured images.  Neural-field methods model the canonical body with NeRF, SDF, occupancy, or learned blend-weight fields~\cite{mildenhall2020nerf,park2019deepsdf,peng2020neural,chen2021snarf,peng2021animatable,athar2022rignerf,weng2022humannerf,yu2023monohuman,jiang2022neuman,jiang2022instantavatar,guo2023vid,jiang2022selfrecon,zheng2023avatarrex}; they are expressive but costly to optimize and must resolve difficult posed-to-canonical correspondences.  Explicit Gaussian and mesh variants improve rendering speed and make LBS-style animation more direct~\cite{xu2024gaussian,jiang2024robust,shao2024splattingavatar,pang2023ash,saito2023relightable,moreau2024human,li2024gaussianbody,zielonka2023drivable,chen2024meshavatar,hu2023gauhuman,ye2023animatable,liu2023animatable,lei2023gart,kocabas2023hugs,qian2023dgsavatar,hu2023gaussianavatar,li2024animatable,wen2024gomavatar,liu2025creating,tan2024dressrecon}. ExAvatar~\cite{moon2024expressive} uses SMPL-aware regularization and triplane-parameterized~\cite{chan2021efficient} Gaussian attributes, but still requires subject-specific training that ranges from tens of minutes to hours.

\vspace{-0.1in}
\paragraph{Generative and native 3D avatar methods.}
Generative avatar methods reduce capture requirements by hallucinating missing views. Some optimize a representation with SDS~\cite{poole2022dreamfusion} or related generative priors~\cite{tang2023dreamgaussian,cao2023dreamavatar,liu2023humangaussian,zhang2023humanref,weng2023zeroavatar,huang2024dreamwaltzg}.
Others use multiview or video diffusion for consistency or refinement~\cite{shi2023zero,shi2023mvdream,liu2023syncdreamer,szymanowicz2023viewset,taubner2024cap,bhler2025dream,huang2025adahuman,liu2024humanvdm,zhou2025zero,qiu2024anigs,cha2024perse,wang2025humandreamerx,guo2025vid,lu2025gas}.
These systems can operate from a single image, but they consume even more compute than optimization-based fitting methods.
Native 3D and large reconstruction models~\cite{wang2022rodin,zhang2024rodinhd,zhang2024clay,zhao2025hunyuan,lai2025hunyuan,li2025triposg,yang2025fast,xiang2026rng} learn 3D distributions over massive 3D datasets~\cite{deitke2022objaverse,deitke2023objaversexl}.
Most of these methods require massive 3D assets and have difficulty scaling to the human domain due to the lack of large-scale, high-quality 3D data. SIGMAN~\cite{yang2025sigman} is the closest native 3D human baseline, and we compare against it.

\vspace{-0.1in}
\paragraph{Feedforward avatar reconstruction.}
Earlier feedforward human methods~\cite{kanazawa2018endtoend,kolotouros2019learning,kocabas2019vibe,li2022cliff,zhang2021pymaf,cai2023smplerx} made avatar reconstruction efficient but typically produced static meshes or volumetric implicits rather than real-time animatable avatars~\cite{saito2019pifu,saito2020pifuhd,huang2020arch,he2021arch,zheng2019deephuman,zhang2023sifu,song2023difu,zheng2020pamir,xiu2021icon,xiu2022econ,liao2023highfidelity,zhang2023globalcorrelated}. Generalizable human renderers and earlier animatable models introduced canonical body spaces and learned view synthesis~\cite{wang2022neural,gao2022mpsnerf,hu2023sherf,yang2023human}, setting the stage for recent scaled feedforward systems.
IDOL~\cite{zhuang2024idol} regresses UV-space Gaussians with a Sapiens-scale encoder; LHM~\cite{qiu2025lhm} uses a large transformer trained on 300K+ private videos; LHM++~\cite{qiu2025lhmb} uses Point Transformer~\cite{wu2023point} and extends it to multiple images.
LCA~\cite{li2026largescale} pushes scaled-up training to an extreme with massive internal data.
ARG-Avatar belongs to this feedforward avatar category, but instead of scaling up model and data, it revisits the geometry of template-based avatar reconstruction, resulting in a strong model with substantially fewer trainable parameters.

\vspace{-0.1in}
\paragraph{Geometry-aware attention.}
More recent literature explores injecting geometric guidance into attention with positional encodings and rotary position embeddings (RoPE)~\cite{su2021roformer}.
GTA~\cite{miyato2023gta} generalizes rotary encodings from 1D sequences to relative camera transforms on multiview tokens, and PRoPE~\cite{li2025cameras} and RayRoPE~\cite{wu2026rayrope} further use projective frusta or ray segments to guide RoPE attention, an idea carried into stereo and multiview video diffusion~\cite{sun2026stereo,fan2025omniview}.
We propose FACRoPE with two distinct novelties: first, Foreground Avatar Coordinates (FAC) link tokens in 2D animated and rendered space with 3D canonical tokens through an articulated human template; second, a rendering-in-the-loop pipeline guides attention by decoding and rendering an avatar from intermediate tokens. 

\begin{figure*}[t]
    \centering
    \includegraphics[width=0.95\textwidth]{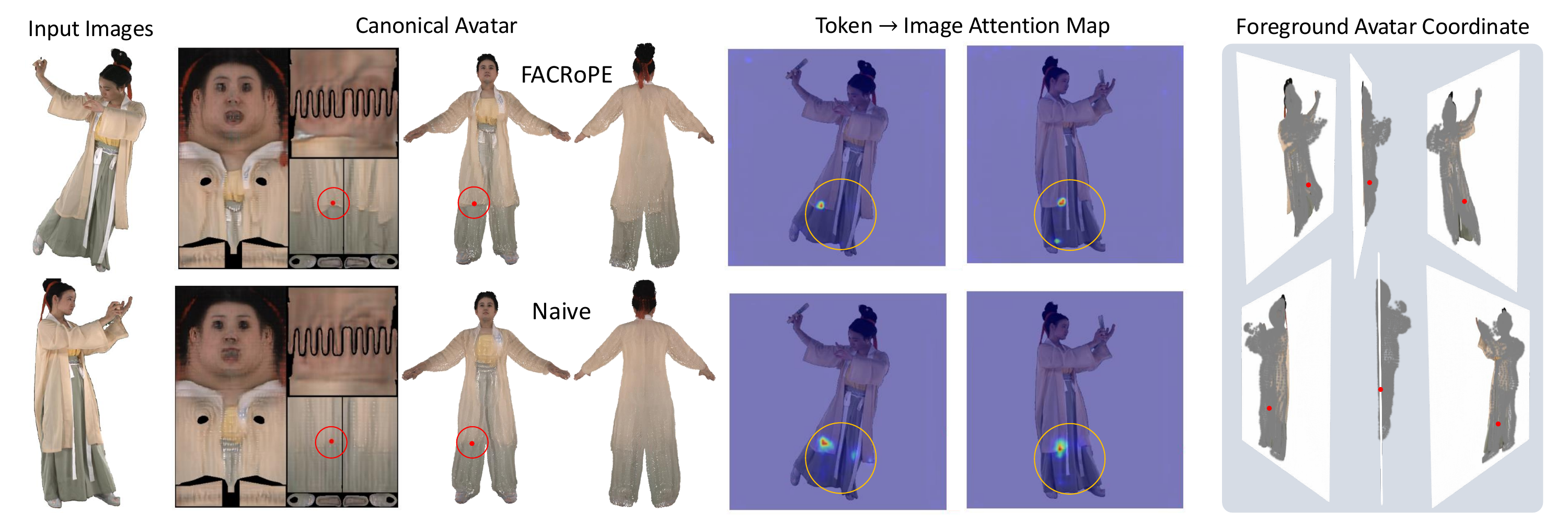}
    \vspace{-0.15in}
    \caption{Comparison of token-to-image attention using FACRoPE (\cref{sec:method:mvargblock}) and naive attention. From left to right: input images, the reconstructed canonical UV 3DGS avatar, the marked token's attention over image features, and FAC visualizations in three views.}
    \vspace{-0.15in}
    \label{fig:fac_attn}
\end{figure*}

\section{Method}
\label{sec:method}

Given a set of input observations $\mathcal{I}=\{\Iv_v\}_{v=1}^{V}$, ranging from casual photos to monocular video frames, together with a set of desired output poses $\mathcal{P}=\{\theta_t\}_{t=1}^{T}$ (skeletal joint angles), we aim to reconstruct a canonical avatar $\Cm$ bound to a template and a pose-dependent avatar $\Gm_t$ for each target pose.

\subsection{Avatar Representation}
\label{sec:method:avatar_representation}

\paragraph{3DGS.}
We use 3D Gaussian Splatting (3DGS)~\cite{kerbl2023gaussian} to represent the avatar. The 3D object is expressed as a set of 3D Gaussian blobs, each parameterized by $[\muv, \qv, \sv, \alpha, \shv]$, where $\muv \in \mathbb{R}^{3}$ is the center, $\qv \in \mathbb{R}^{4}$ is the rotation quaternion, $\sv \in \mathbb{R}^{3}$ is the scale, $\alpha \in (0,1)$ is the opacity, and $\shv \in \mathbb{R}^{(K+1)^2 \times 3}$ contains the spherical-harmonic coefficients up to degree $K$ for view-dependent color $\cv_i(\mathbf{d})$. Projected primitives are splatted, sorted, and alpha-composited:
\begin{equation}
    C(\xv) = \sum_{i=1}^M T_i (\xv) \,\alpha_i(\xv)\,\cv_i(\mathbf{d}), \ 
    T_i(\xv) = \prod_{j < i} \bigl(1 - \alpha_j(\xv)\bigr),
\label{eq:3dgs_rendering}
\end{equation}
where $M$ is the number of primitives contributing to pixel $\xv$, and $\alpha_i(\xv)$ is the projected opacity.

\vspace{-0.1in}
\paragraph{Canonical UV 3DGS Avatar.}\label{sec:method:canonical_uv_3dgs_avatar}
We represent avatars with 3D Gaussians anchored on the UV map of a human mesh template, MHR~\cite{ferguson2025mhr}.
MHR consists of a rest-pose body mesh $\mathcal{M}_0=(\mathcal{V}_0,\mathcal{F})$ with vertices $\mathcal{V}_0=\{\mathbf{v}_i\}_{i}$ and faces $\mathcal{F}$.
The shape of the mesh is presented in Appendix~\ref{app:rest_pose_mesh}.
Its UV map is an atlas $\Omega\subset[0,1]^2$ in which each UV coordinate is associated with a face index $f_i \in \mathcal{F}$ and a set of barycentric weights $\boldsymbol{\beta}_i\in\mathbb{R}^3$.
We place one Gaussian at each UV grid point; its rest-pose anchor $\mathbf{a}_i\in\mathbb{R}^{3}$ is the barycentric interpolation of the vertices of face $f_i$.
The position of each Gaussian is its anchor plus a learned canonical offset, $\muv_i=\mathbf{a}_i+\mathbf{o}_i$, where $\ov_i\in\mathbb{R}^{3}$.
We define the canonical avatar as
\begin{equation}
    \Cm = \{(\ov_i,\qv_i,\sv_i,\alpha_i,\shv_i)\}_{i=1}^{L},
    \label{eq:canonical_avatar}
\end{equation}
where $L$ is the total number of Gaussian primitives.
We also employ a customized UV rasterization strategy to allocate higher resolution to some body parts, resulting in $L=264K$ Gaussians (see Appendix~\ref{app:adaptive_uv_resolution}).

\noindent\textbf{Pose-Deformed Avatar.}\label{sec:method:pose_deformed_avatar}
In real-world images, subjects exhibit non-rigid deformations due to loose clothing, long hair, \etc. To explain these effects, we predict a pose-dependent deformation $\mathbf{d}_{t,i}\in\mathbb{R}^{10}$ for each primitive under target pose $t$: $\delta \mathbf{d}_{t,i}=[\delta\mathbf{o}_{t,i},\delta\sv_{t,i},\delta\qv_{t,i}]$, where $\delta\mathbf{o}_{t,i}\in\mathbb{R}^{3}$, $\delta\sv_{t,i}\in\mathbb{R}^{3}$, and $\delta\qv_{t,i}\in\mathbb{R}^{4}$ adjust the position, scale, and orientation of the primitive, respectively.
The pose-dependent residuals adjust the canonical avatar before animation:
\begin{equation}
    \begin{aligned}
    \Cm_t &= \{(\muv_{t,i}, \qv_i \otimes \delta\qv_{t,i}, \sv_i \cdot \delta\sv_{t,i}, \alpha_i,\shv_i)\}_{i=1}^{L}, \\
    \muv_{t,i} &= \mathbf{a}_i+\mathbf{o}_i + \delta \mathbf{o}_{t,i},
    \end{aligned}
    \label{eq:posed_avatar}
\end{equation}
where $\otimes$ denotes quaternion multiplication.

\begin{algorithm}[t]
    \caption{Foreground Avatar Coordinate (FAC)}
    \label{alg:fac}
    \begin{algorithmic}[1]
    \Require UV tokens $\xv \in \mathbb{R}^{L\times D}$, UV patches $\{\mathcal{P}_i\}$, image pose $\thetav_v$, camera matrix $\Pv_v \in \mathbb{R}^{3 \times 4}$, candidate avatar $\Cm$
    \Ensure $\mathrm{FAC} \in \mathbb{R}^{L \times 3}$
    \State $\Gm_v = [\pv, \ldots] \gets \mathrm{LBS}(\Cm,\,\thetav_v)$ \Comment{animated avatar}
    \State $\muv_i \gets \dfrac{1}{|\mathcal{P}_i|}\displaystyle\sum_{j \in \mathcal{P}_i} \pv_j$ \Comment{animated UV token position}
    \State $D_v^{\mathrm{map}} \gets \mathcal{R}_d(\Gm_v;\,\Pv_v)$ \Comment{render depth map}
    \State $[\xv,\,\yv,\,\zv] \gets \Pv_v\,\muv$ \Comment{token positions in camera coordinates}
    \State $\Dv \gets \mathrm{sample}(D_v^{\mathrm{map}},\,\xv,\,\yv)$ \Comment{depth vector}
    \State \Return $\bigl[\xv,\;\yv,\;\zv - \Dv\bigr]$
    \end{algorithmic}
\end{algorithm}

\begin{figure*}[t]
    \centering
    \begin{nolinenumbers}
    \noindent\makebox[\textwidth][c]{%
        \hspace*{-36pt}\resizebox{1\textwidth}{!}{\input{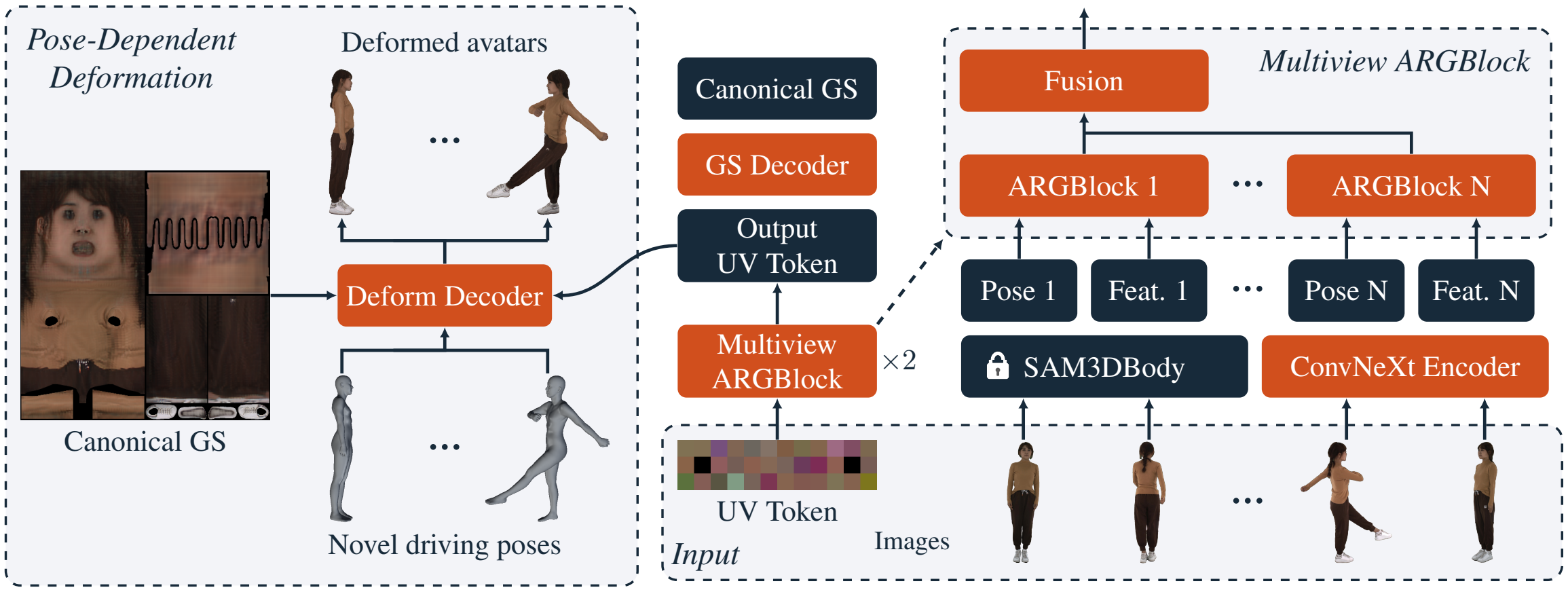}}\hspace*{36pt}%
    }%
    \end{nolinenumbers}
    \vspace{-0.1in}
    \caption{ARG-Avatar network pipeline. Each input image is processed by SAM3DBody and a ConvNeXt encoder to obtain its pose and features. ARG-Avatar then applies two MV-ARGBlocks to attend to and fuse features from multiple input images. The output tokens are decoded into a canonical Gaussian avatar and pose-dependent deformations for each driving pose.}
    \label{fig:ARG-Avatar_overview}
    \vspace{-0.15in}
\end{figure*}

\vspace{-0.1in}
\paragraph{Forward LBS.}
\label{sec:method:forward_lbs}
We animate the avatar using the template mesh and linear blend skinning (LBS)~\cite{loper2023smpl}. The skeleton $\mathcal{B}$ consists of rest-pose joints $\{\mathbf{r}_j\}_{j=1}^{J}$, pose parameters $\theta_t=\{\theta_{t,j}\}_{j=1}^{J}$, and a kinematic-tree parent map $\pi(j)$, together with per-vertex skinning weights $W^m=[w^m_{i,j}]\in\mathbb{R}^{N_m\times J}$ that bind each vertex to the joints.
Accumulating the per-joint local transforms along the kinematic tree yields the global joint transforms $\mathbf{T}_{t,j}$ at pose $t$ and $\mathbf{T}_{c,j}$ in the rest pose.
We extend LBS to operate on the UV 3DGS avatar.
Since our primitives are indexed by UV coordinates, we barycentrically interpolate the per-vertex transforms at each Gaussian's anchor, yielding the position $\tilde{\mathbf{a}}_{t,k}$ and rotation transform $\qv^{\mathrm{rigid}}_{t,k}$.
Formally, our LBS transformation is defined as follows:
\begin{equation}
    \begin{aligned}
        \Gm_t &= \mathrm{LBS}_t(\Cm; \theta_t) \\
        & = \{(\mathbf{p}_k, \qv^{\mathrm{rigid}}_{t,k} \otimes \qv_k, \sv_k, \alpha_k, \shv_k)\}_{k=1}^{L}, \\
        \mathbf{p}_k &= \tilde{\mathbf{a}}_{t,k} + \qv^{\mathrm{rigid}}_{t,k} \cdot \mathbf{o}_k.
    \end{aligned}
    \label{eq:lbs_operator}
\end{equation}
For the pose-deformed avatar $\Cm_t$, we replace the position, orientation, and scale with their deformed counterparts from Eq.~\ref{eq:posed_avatar}.
A more detailed description of UV 3DGS animation is provided in Appendix~\ref{app:lbs_details}.

\subsection{Guiding Attention with Avatar Geometry}
\label{sec:method:foreground_avatar_coordinate}

Template-based feedforward avatar reconstruction requires extracting appearance and geometry information from the input images and transferring it to the corresponding locations on a predefined avatar template. 

Consider the bottom panel of~\cref{fig:fac_attn}. Avatar token $i$, marked by the red dot and circle on the UV 3DGS map, corresponds to the \textbf{edge} of the left leg.
With naive cross-attention between the canonical avatar and the input image, each avatar token must search over the entire image and learn the correspondence between template locations and image regions solely from training data. Although absolute positional embeddings are added to the avatar tokens, learning these correspondences remains challenging.


This limitation is illustrated in Row 2 of ``Token $\rightarrow$ Image Attention Map'' in~\cref{fig:fac_attn}. Naive attention covers a broad, diffuse image region, as shown in the yellow circle. Moreover, some attention is assigned to an incorrect location outside the clothing boundary. Consequently, the avatar reconstructed with naive attention, shown in Row 2, Column 3, is blurry and lacks fine-grained details.

Our key observation is that the image location corresponding to each avatar element is largely determined by the known avatar template and camera geometry. Therefore, this correspondence does not need to be learned entirely from data. Motivated by this observation, we use the avatar template to explicitly guide the attention mechanism toward the appropriate image regions.

\vspace{-0.1in}
\paragraph{Coordinate-based guidance.}
We use Rotary Position Embedding (RoPE)~\cite{su2021roformer} as a framework for injecting geometric guidance.
For queries and keys with associated positions, RoPE makes attention scores depend on their relative coordinate offsets.
Given queries $\Qv$ and keys $\Kv$ carrying coordinates $\cv^q$ and $\cv^k$, RoPE rotates each token according to its coordinates before computing the attention product:
\begin{equation}
    \mathrm{RoPEAttn}(\Qv,\Kv,\Vv) = \mathrm{Attn}(\Rv(\cv^q) \Qv,\Rv(\cv^k) \Kv,\Vv),
    \label{eq:rope_attention}
\end{equation}
where $\Rv(\cdot)$ applies a per-token block-diagonal rotation built from that token's coordinates.
Consequently, the similarity between query token $i$ and key token $j$ depends on the relative coordinate offset through the relative rotation $\Rv(\cv_j^k-\cv_i^q)$.
In our cross-attention, UV tokens serve as queries and image tokens as keys. Aligning their coordinates helps each UV token match its corresponding image token.

\vspace{-0.1in}
\paragraph{Foreground Avatar Coordinate.}
We first assign each image token its image-plane coordinates, $(u, v, 0)$, where $u$ and $v$ are both in $[0, 1]$.
These coordinates serve as $\cv_k$ in RoPE attention.
We then assign each UV token the coordinates of its projected image location after animation and rendering.
The algorithm is described in~\cref{alg:fac}.
The resulting coordinate is referred to as FAC and serves as $\cv_q$ in~\cref{eq:rope_attention}.
Given a candidate avatar $\Cm^{\mathrm{cand}}$, we animate it to the pose observed in image $v$, project it into the camera view, and render its depth.
The candidate avatar can be a neutral mesh or a coarse Gaussian avatar obtained through Intermediate Token Rendering (ITR, \cref{sec:method:intermediate_token_rendering}).

As shown in Row 1 of~\cref{fig:fac_attn}, FAC makes the attention precise and sharp.
The last column of~\cref{fig:fac_attn} visualizes the coordinates relative to the image plane: the coordinates of visible tokens lie on the plane, whereas those of occluded tokens are pushed away from it.

\begin{figure*}[h]
    \centering
    \noindent\makebox[\textwidth][c]{%
        \hspace*{-21pt}\resizebox{1\textwidth}{!}{\input{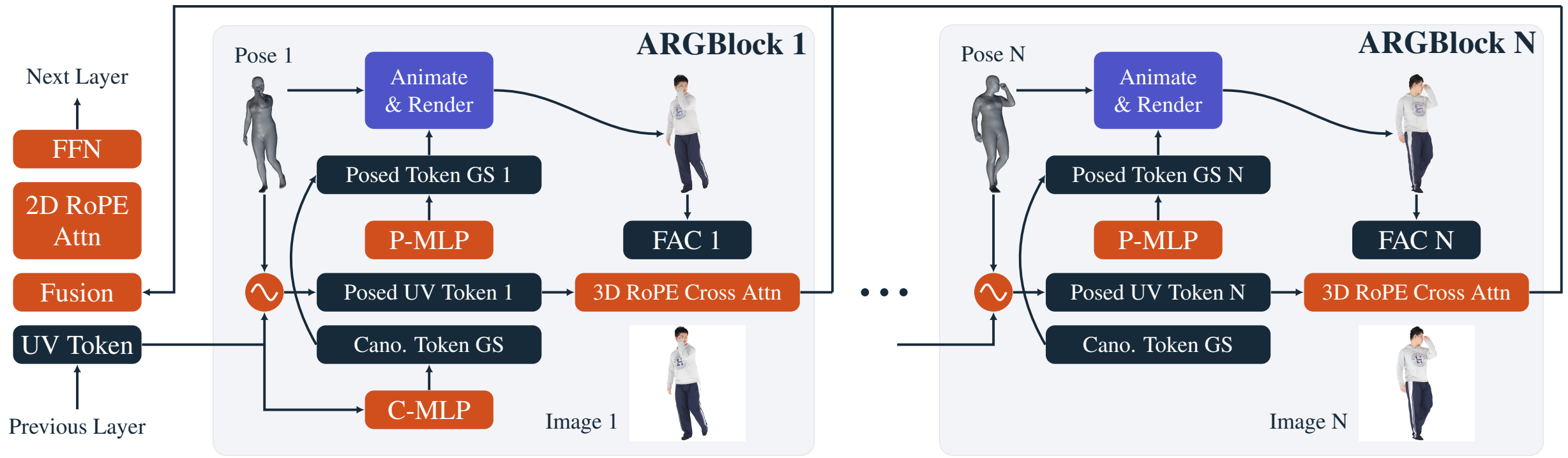}}\hspace*{21pt}%
    }
    \vspace{-0.25in}
    \caption{Pipeline of the Animation-Rendering Guidance Block (ARGBlock) with Intermediate Token Rendering (ITR). UV tokens are decoded into candidate Gaussian avatars, which are then animated and rendered to obtain FAC that guides attention to image features.}
    \label{fig:mvargblock}
    \vspace{-0.15in}
\end{figure*}

\subsection{ARG-Avatar Network}
\label{sec:method:architecture}

The pipeline of the ARG-Avatar network is shown in \cref{fig:ARG-Avatar_overview}.

\noindent ARG-Avatar $F(\bar{\Cm}, \{\Iv_v\}_{v=1}^{V})$ takes as input a zero-initialized 3DGS UV map and $V$ images.
The UV map is patchified into $N_T$ UV tokens $\xv \in \mathbb{R}^{N_T\times D}$, which are refined through $M=2$ multiview ARGBlocks.
The output UV tokens are decoded into a canonical avatar $\Cm$ and pose-dependent deformations for each target driving pose $\thetav$.

\vspace{-0.1in}
\paragraph{Feature Extraction.}
Each input image is processed through a frozen SAM3DBody~\cite{yang2026sam} model and a trainable ConvNeXt~\cite{woo2023convnext} model.
SAM3DBody estimates a per-image body pose $\thetav_v$ and supplies dense ViT backbone features~\cite{oquab2023dinov} that are added to the ConvNeXt features, forming per-view image tokens $F_v \in \mathbb{R}^{H_f \times W_f \times D_f}$.

\vspace{-0.1in}
\paragraph{Multiview ARGBlock with ITR.}
\label{sec:method:mvargblock}
\label{sec:method:intermediate_token_rendering}
The block architecture with intermediate token rendering is shown in~\cref{fig:mvargblock}.
First, the UV tokens $\xv$ are cloned into $V$ branches, $\{\xv^1, \ldots, \xv^V\}$, one for each input image.
They are augmented with pose-specific embeddings, becoming posed UV tokens.
Second, we perform ITR to obtain the candidate avatar used to compute FAC.
A simple MLP, CMLP, decodes one Gaussian from each intermediate token.
A second MLP, PMLP, decodes the pose-dependent deformation for each Gaussian.
Following~\cref{eq:posed_avatar}, we combine these predictions into the posed token-GS avatar and pass it to~\cref{alg:fac} to obtain $\mathrm{FAC}_v$.
The UV tokens then cross-attend to image features using 3D RoPE guided by $\mathrm{FAC}_v$.
Finally, the output features from each pose branch are fused into the main UV token stream and pass through another self-attention layer and a feedforward network (FFN).
The fusion uses elementwise attention, in which each main UV token $\xv_i \in \mathbb{R}^{1 \times D}$ attends to its corresponding $V$ tokens, $[\xv_i^1, \ldots, \xv_i^{V}] \in \mathbb{R}^{V \times D}$.


\vspace{-0.1in}
\paragraph{Gaussian Decoder.}
The output UV tokens are decoded into a canonical avatar and pose-dependent deformations for each target pose.
We first rearrange the UV tokens on the UV map to restore its 2D layout.
Then, a CNN decoder $\mathrm{CDec}$ upsamples the token map to the output resolution.
The pose-dependent deformations are predicted with another CNN, $\mathrm{PDec}$, with a similar structure.

\subsection{Training}

ARG-Avatar is trained with photometric losses to reconstruct input and driving-pose images, following other avatar reconstruction methods~\cite{zhuang2024idol, qiu2025lhmb, qiu2025lhm}.
Here, we introduce the proposed regularizer; the complete loss formulation is provided in Appendix~\ref{app:loss_details}.

\vspace{-0.1in}
\paragraph{Common Shell Transfer.}
We found that the posed deformation tends to disentangle from the deformation of the canonical avatar.
To regularize this, we encourage the per-view pose offsets to have no shared component; any shared component should instead be captured by the canonical offset.
In other words, the canonical offset should explain deformation patterns common to all input images.

We call the proposed regularizer the common shell-transfer regularizer.
It first estimates a robust shared \emph{shell}, $\bar{\delta\ov} \in \mathbb{R}^{L \times 3}$, across the $V$ per-view pose offsets $\{\delta\ov_{v,i}\}_{v=1}^{V}$:
\vspace{-0.1in}
\begin{equation}
    \begin{aligned}
    \bar{\delta\ov}_i &= \frac{1}{\sum_{v} w_{v,i}} \sum_{v=1}^{V} w_{v,i}\,\delta\ov_{v,i}, \\
    w_{v,i} &= \left(\|\delta\ov_{v,i} - \bar{\delta\ov}_i^{(0)}\|_2^2 + \varepsilon^2 \right)^{-\frac{1}{2}},
    \end{aligned}
    \label{eq:shell_mean}
\end{equation}
where $\bar{\delta\ov}_i^{(0)}$ is an unweighted initialization and $\varepsilon=10^{-3}$.
The common shell-transfer loss is then computed as
\begin{equation}
    \begin{aligned}
    \mathcal{L}_{\text{shell}} &=
    \sum_i c_i\,\bigl\|\ov_i - \mathrm{sg}(\ov_i + \bar{\delta\ov}_i)\bigr\|_2^2 \\
    &\quad + \frac{1}{V}\sum_{v,i} c_i\,\bigl\|\delta\ov_{v,i} - \mathrm{sg}(\delta\ov_{v,i} - \bar{\delta\ov}_i)\bigr\|_2^2,
    \end{aligned}
    \label{eq:shell_loss}
\end{equation}
where $c_i = \min(1, \frac{\varepsilon}{V} \sum_{v=1}^{V} w_{v,i}) \in [0,1]$ is the confidence weight and $\mathrm{sg}(\cdot)$ is the stop-gradient operation.

\section{Experiments}
\label{sec:experiments}

\begin{table*}[t]
    \centering
    \caption{Comparison of ARG-Avatar with the baselines. We report novel-pose synthesis on video datasets, novel-view synthesis on 3D scans, and input-pose reconstruction on NeuMan. ``Multiview Videos'' combines DNARendering and 4D-Dress; ``3D Scans'' averages results across 2K2K, THuman2.1, and Human4DiT. NeuMan is unseen during training. Other datasets use validation splits constructed according to~\cref{sec:experiments:datasets}. \protect\colorbox[rgb]{0.957,0.490,0.459}{First}, \protect\colorbox[rgb]{1.000,0.922,0.400}{second}, and \protect\colorbox[rgb]{1.000,0.949,0.800}{third} indicate the top three results in each metric column.}
    \label{tab:main_results}
    \vspace{-0.1in}
    {\scriptsize
    \setlength{\tabcolsep}{2pt}
    \setlength{\arrayrulewidth}{0.3pt}
    \renewcommand{\arraystretch}{1.0}
    \resizebox{\textwidth}{!}{%
    \begin{tabular}{>{\raggedright\arraybackslash}p{1.7cm}l|cccccc|ccc|ccc|ccc}
        \whline{1.2pt}
        & & \multicolumn{6}{c|}{\textit{NeuMan}} & \multicolumn{3}{c|}{\textit{Multiview Videos}} & \multicolumn{3}{c|}{\textit{3D Scans}} & \multicolumn{3}{c}{\textit{HumanVid}} \\
        Method & $V$ & \multicolumn{3}{c}{Novel Poses} & \multicolumn{3}{c|}{Input Poses} & \multicolumn{3}{c|}{Novel Poses} & \multicolumn{3}{c|}{Novel Views} & \multicolumn{3}{c}{Novel Poses} \\
         &  & PSNR $\uparrow$ & SSIM $\uparrow$ & LPIPS $\downarrow$ & PSNR & SSIM & LPIPS & PSNR & SSIM & LPIPS & PSNR & SSIM & LPIPS & PSNR & SSIM & LPIPS \\
        \hline
        \parbox[c]{1.7cm}{\raggedright SIGMAN~\cite{yang2025sigman}} & 1 & 19.62 & 0.957 & 0.066 & 19.96 & 0.959 & 0.063 & 16.52 & 0.944 & 0.076 & 15.27 & 0.908 & 0.112 & 16.17 & 0.862 & 0.162 \\
        \whline{0.6pt}
        \parbox[c]{1.7cm}{\raggedright IDOL~\cite{zhuang2024idol}} & 1 & 24.96 & 0.964 & 0.036 & 25.35 & 0.966 & 0.033 & 23.18 & 0.962 & 0.051 & 21.32 & 0.935 & 0.086 & 20.53 & 0.899 & 0.129 \\
        \whline{0.6pt}
        \parbox[c]{1.7cm}{\raggedright LHM~\cite{qiu2025lhm}} & 1 & 26.10 & 0.968 & 0.028 & 28.40 & 0.972 & 0.019 & 25.30 & 0.965 & 0.041 & 22.81 & 0.940 & 0.078 & 21.23 & 0.906 & 0.111 \\
        \whline{0.6pt}
        \multirow[c]{4}{*}{\parbox[c]{1.7cm}{\raggedright LHM++~\cite{qiu2025lhmb}}} & 1 & 27.71 & 0.970 & 0.023 & 30.62 & \cellcolor[rgb]{1.000,0.922,0.400}0.974 & 0.017 & 25.34 & \cellcolor[rgb]{1.000,0.949,0.800}0.966 & 0.039 & 22.31 & 0.939 & 0.078 & 22.03 & 0.910 & 0.104 \\
         & 2 & 27.97 & 0.971 & 0.022 & 28.86 & 0.972 & 0.019 & 25.31 & \cellcolor[rgb]{1.000,0.949,0.800}0.966 & 0.038 & 22.42 & 0.939 & 0.075 & 22.35 & 0.911 & 0.101 \\
         & 4 & 28.03 & \cellcolor[rgb]{1.000,0.949,0.800}0.972 & 0.022 & 28.63 & \cellcolor[rgb]{1.000,0.949,0.800}0.973 & 0.020 & 25.60 & \cellcolor[rgb]{1.000,0.949,0.800}0.966 & 0.037 & 22.57 & 0.940 & 0.074 & 22.47 & 0.911 & 0.099 \\
         & 6 & 28.09 & \cellcolor[rgb]{1.000,0.949,0.800}0.972 & 0.022 & 28.50 & \cellcolor[rgb]{1.000,0.949,0.800}0.973 & 0.020 & 25.68 & \cellcolor[rgb]{1.000,0.922,0.400}0.967 & 0.036 & 22.49 & 0.940 & 0.074 & 22.50 & \cellcolor[rgb]{1.000,0.949,0.800}0.912 & 0.098 \\
        \whline{0.6pt}
        \multirow[c]{4}{*}{\parbox[c]{1.7cm}{\raggedright Ours}} & 1 & 27.02 & 0.966 & 0.024 & 28.67 & \cellcolor[rgb]{1.000,0.922,0.400}0.974 & \cellcolor[rgb]{1.000,0.949,0.800}0.016 & 25.86 & 0.962 & 0.034 & 24.09 & 0.937 & 0.058 & 21.64 & 0.897 & 0.100 \\
         & 2 & \cellcolor[rgb]{1.000,0.949,0.800}29.17 & 0.971 & \cellcolor[rgb]{1.000,0.949,0.800}0.019 & \cellcolor[rgb]{0.957,0.490,0.459}\textbf{33.88} & \cellcolor[rgb]{0.957,0.490,0.459}\textbf{0.981} & \cellcolor[rgb]{0.957,0.490,0.459}\textbf{0.010} & \cellcolor[rgb]{1.000,0.949,0.800}27.53 & 0.965 & \cellcolor[rgb]{1.000,0.949,0.800}0.027 & \cellcolor[rgb]{1.000,0.949,0.800}25.64 & \cellcolor[rgb]{1.000,0.949,0.800}0.941 & \cellcolor[rgb]{1.000,0.949,0.800}0.046 & \cellcolor[rgb]{1.000,0.949,0.800}24.57 & \cellcolor[rgb]{1.000,0.949,0.800}0.912 & \cellcolor[rgb]{1.000,0.949,0.800}0.076 \\
         & 4 & \cellcolor[rgb]{1.000,0.922,0.400}30.14 & \cellcolor[rgb]{1.000,0.922,0.400}0.973 & \cellcolor[rgb]{1.000,0.922,0.400}0.017 & \cellcolor[rgb]{1.000,0.949,0.800}33.26 & \cellcolor[rgb]{0.957,0.490,0.459}\textbf{0.981} & \cellcolor[rgb]{1.000,0.922,0.400}0.011 & \cellcolor[rgb]{1.000,0.922,0.400}28.53 & \cellcolor[rgb]{1.000,0.922,0.400}0.967 & \cellcolor[rgb]{1.000,0.922,0.400}0.024 & \cellcolor[rgb]{1.000,0.922,0.400}26.59 & \cellcolor[rgb]{1.000,0.922,0.400}0.944 & \cellcolor[rgb]{1.000,0.922,0.400}0.040 & \cellcolor[rgb]{1.000,0.922,0.400}25.44 & \cellcolor[rgb]{1.000,0.922,0.400}0.916 & \cellcolor[rgb]{1.000,0.922,0.400}0.069 \\
         & 6 & \cellcolor[rgb]{0.957,0.490,0.459}\textbf{30.82} & \cellcolor[rgb]{0.957,0.490,0.459}\textbf{0.975} & \cellcolor[rgb]{0.957,0.490,0.459}\textbf{0.016} & \cellcolor[rgb]{1.000,0.922,0.400}33.30 & \cellcolor[rgb]{0.957,0.490,0.459}\textbf{0.981} & \cellcolor[rgb]{1.000,0.922,0.400}0.011 & \cellcolor[rgb]{0.957,0.490,0.459}\textbf{28.83} & \cellcolor[rgb]{0.957,0.490,0.459}\textbf{0.968} & \cellcolor[rgb]{0.957,0.490,0.459}\textbf{0.023} & \cellcolor[rgb]{0.957,0.490,0.459}\textbf{26.81} & \cellcolor[rgb]{0.957,0.490,0.459}\textbf{0.945} & \cellcolor[rgb]{0.957,0.490,0.459}\textbf{0.039} & \cellcolor[rgb]{0.957,0.490,0.459}\textbf{25.86} & \cellcolor[rgb]{0.957,0.490,0.459}\textbf{0.918} & \cellcolor[rgb]{0.957,0.490,0.459}\textbf{0.067} \\
        \whline{1.2pt}
    \end{tabular}
    }
    }
    \vspace{-0.1in}
\end{table*}

\subsection{Experimental Setup}
\label{sec:experiments:protocol}

\paragraph{Datasets.}\label{sec:experiments:datasets}
We train ARG-Avatar using only publicly available datasets: DNARendering~\cite{cheng2023dnarendering}, 2K2K~\cite{han2023highfidelity}, 4D-Dress~\cite{wang2024ddress}, Human4DiT~\cite{shao2024human}, THuman2.1~\cite{yu2021function}, and HumanVid~\cite{wang2024humanvid}. Together, these datasets contain 6.8K subjects and 64K sequences.

We evaluate the model on NeuMan~\cite{jiang2022neuman}, a standard benchmark for evaluating animatable avatars, and on validation splits of the training datasets. We construct the validation splits by randomly selecting $10\%$ of the subjects in each dataset and excluding them from training.

\vspace{-0.1in}
\paragraph{Training.}
ARG-Avatar is trained for 100K steps with a global batch size of 96 using AdamW~\cite{loshchilov2017decoupled}.
We use a learning rate of $10^{-4}$ with cosine scheduling.
Training takes around 200 hours on four A6000 GPUs.
For the ablation study, we train all variants at $512^2$ UV resolution for 50K steps and observe that their training losses have converged.
Full training details are available in Appendix~\ref{app:training_details}.

\vspace{-0.1in}
\paragraph{Baselines and metrics.}\label{sec:experiments:baselines}
We compare ARG-Avatar with four feedforward baselines trained on large-scale datasets: SIGMAN~\cite{yang2025sigman}, IDOL~\cite{zhuang2024idol}, LHM~\cite{qiu2025lhm}, and LHM++~\cite{qiu2025lhmb}. We exclude LCA~\cite{li2026largescale} because, at the time of submission, it had neither reported results on public benchmarks nor released model checkpoints.

We evaluate all methods on the novel-pose synthesis (NPS) task. Given one or more input views, each method renders the reconstructed avatar in novel target poses, and the resulting images are compared with the corresponding ground-truth images. Following standard practice, we report PSNR, SSIM~\cite{wang2004image}, and LPIPS~\cite{zhang2018unreasonable}.
For both input and target poses on all evaluation datasets, we use MHR poses estimated by SAM3DBody for ARG-Avatar and SMPL-X poses estimated by LHM-Track~\cite{qiu2025lhm} for the baselines. On NeuMan, our evaluation of LHM++ yields 27.97 dB and 28.03 dB with two and four views, respectively, closely matching the 27.83 dB and 27.94 dB reported in the original LHM++ paper~\cite{qiu2025lhmb}. This close agreement supports the reliability of our tracking results for evaluating baseline performance.


\begin{figure*}[t]
    \centering
    \includegraphics[width=0.98\textwidth]{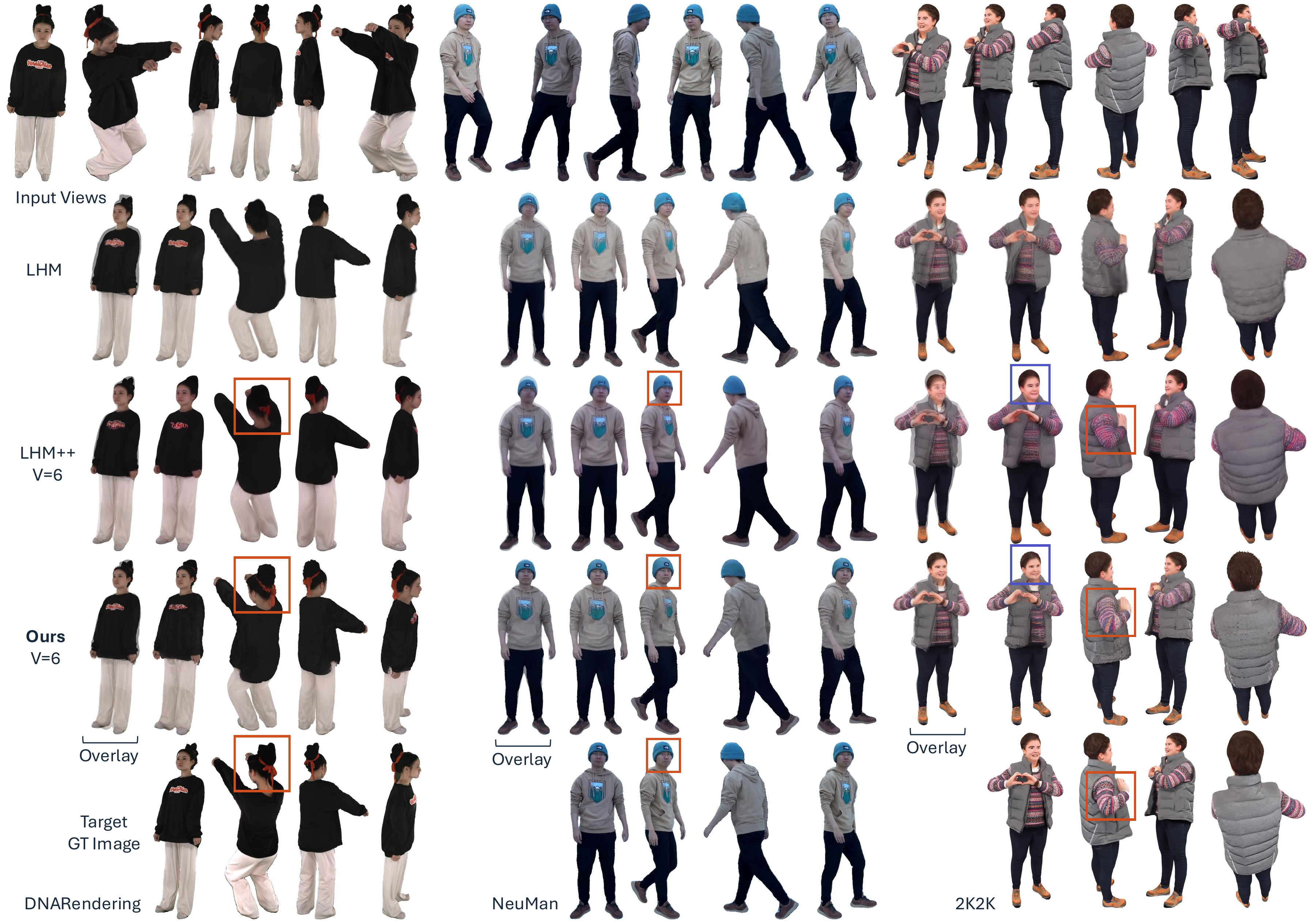}
    \vspace{-0.1in}
    \caption{Qualitative comparison of ARG-Avatar with the baselines. All samples are unseen during ARG-Avatar training, but 2K2K is \textbf{used to train} LHM and LHM++. LHM uses the first view as input; LHM++ and ARG-Avatar use all six views. The first column of each group shows an overlay of the rendered and target images. Orange and blue boxes highlight differences.}
    \label{fig:qualitative}
    \vspace{-0.15in}
\end{figure*}

\subsection{Results}
\label{sec:experiments:results}

\paragraph{Quantitative Results.}
We report the quantitative results for all methods in~\cref{tab:main_results}. Two main findings emerge.

First, ARG-Avatar outperforms all baselines on every dataset with two or more input views. 
With six views ($V=6$), it beats LHM++, the strongest baseline, by 2.7 dB in PSNR on NeuMan, with similar gains on multiview videos (3.2 dB), 3D scans (4.3 dB), and HumanVid (3.4 dB).
Second, ARG-Avatar makes more effective use of additional input views. In the ``input poses'' evaluation, it reaches 33.30 dB with six views, outperforming LHM++ by 4.8 dB. On NeuMan, increasing the number of input views from one to six improves ARG-Avatar by 3.8 dB, whereas LHM++ improves by only 0.4 dB. This indicates that ARG-Avatar integrates information from multiple observations more effectively.


We note that ARG-Avatar is not trained to inpaint invisible regions, so its weaker performance at $V=1$ is expected.
In real-world reconstruction scenarios, users can easily capture multiple input images by recording a video with a smartphone.
The detailed per-dataset quantitative results are presented in Appendix~\ref{app:full_results}.

\vspace{-0.1in}
\paragraph{Qualitative Results.}
We present a qualitative comparison in~\cref{fig:qualitative}.
ARG-Avatar preserves both input texture and geometry more faithfully than the baselines.
First, the orange boxes highlight texture differences.
In the DNARendering example, the red hair band is missing in LHM and only partially reconstructed in LHM++, while ARG-Avatar produces the most complete reconstruction.
Similarly, ARG-Avatar recovers more fine-grained details, such as the logo on the hat in the NeuMan sample and the colorful stripes on the arms in the 2K2K sample.
Second, the blue boxes show the geometric differences.
All three examples show clear shadows around the body in the overlays of the baseline results, while ARG-Avatar has the smallest shadow area.
We note that our SMPL-X tracking is reliable (see~\cref{sec:experiments:baselines}).
In the 2K2K example, the feet are aligned correctly, whereas the head region shows shadows caused by misalignment.

\vspace{-0.1in}
\paragraph{Model Efficiency.}
With only 68M trainable parameters, ARG-Avatar has 6\% of LHM's and 11\% of LHM++'s trainable parameter count. It is trained exclusively on public datasets, using $\sim 21\%$ as much training data as LHM. Training requires around 125 H100-equivalent GPU-hours, 7\% of LHM++'s compute (Appendix~\ref{app:scale_training_compute}). Six-view feedforward reconstruction, including the SAM3DBody forward pass, takes $2.3$ seconds: $6.2\times$ faster than LHM++ ($14.2$ seconds) with comparable GPU memory usage.
Appendix~\ref{app:runtime_profile} provides detailed model profiles.

\subsection{Ablation Study}
\label{sec:experiments:ablations}

We ablate the three key components of ARG-Avatar: FACRoPE, the ITR pipeline, and pose-dependent deformation modeling (``Deform'').
FACRoPE is replaced by vanilla cross-attention with absolute positional embeddings.
ITR is removed by dropping the decoding pathways through CMLP and PMLP in~\cref{fig:mvargblock} and using the template mesh to compute FAC.
Figure~\ref{fig:argblock_variants} in the appendix illustrates these variants.
Pose-dependent deformation is removed by dropping the deformation decoder entirely, leaving only a canonical avatar for animation.

\begin{figure}[t]
    \centering
    \includegraphics[width=0.9\columnwidth]{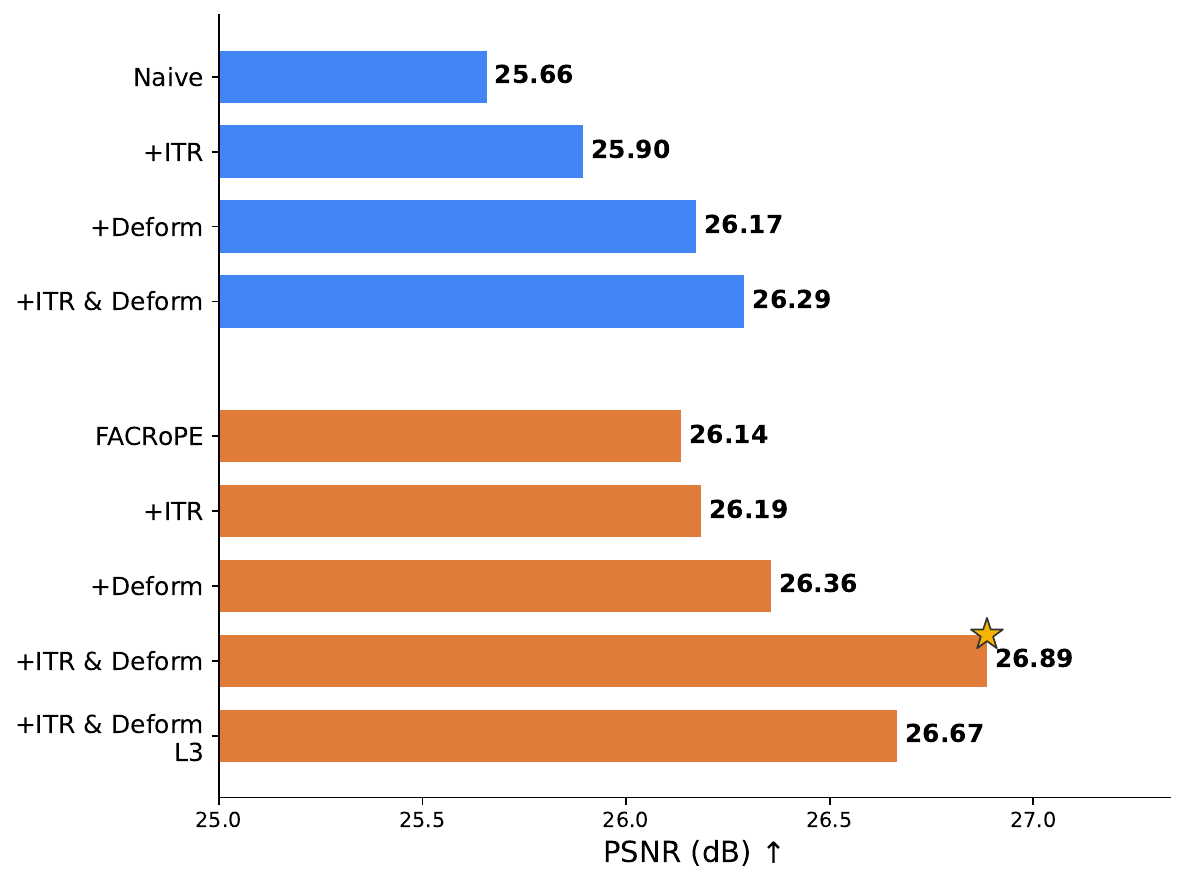}
    \vspace{-0.15in}
    \caption{Ablation study of ARG-Avatar's three key components: FACRoPE attention, the ITR pipeline, and pose-dependent deformation (``Deform''). ``L3'' denotes a variant with one additional layer. The star marks the variant selected for the final model.}
    \vspace{-0.1in}
    \label{fig:ablations}
\end{figure}

Enumerating all $2^3=8$ combinations yields the results in~\cref{fig:ablations}.
First, FACRoPE improves PSNR in every configuration, even with a neutral avatar, demonstrating useful guidance from inaccurate candidates.
Second, ITR and Deform each improve the vanilla-attention variants.
Third, FACRoPE yields smaller gains with either ITR or Deform alone, but substantially larger gains with both.
Finally, three layers do not improve PSNR, so the final model uses two.
In conclusion, all three components are crucial.

Appendix~\ref{app:sam3dbody_ablation} shows that substituting DINOv2 features, as used in LHM++, yields similar performance.

\clearpage
\makeatletter
\global\@colroom\textheight
\global\vsize\textheight
\makeatother
{
    \small
    
    \bibliography{refs}
}

\clearpage
\appendix
\section{Rest-Pose Mesh}
\label{app:rest_pose_mesh}

\begin{figure}[h]
    \centering
    \includegraphics[width=0.49\linewidth]{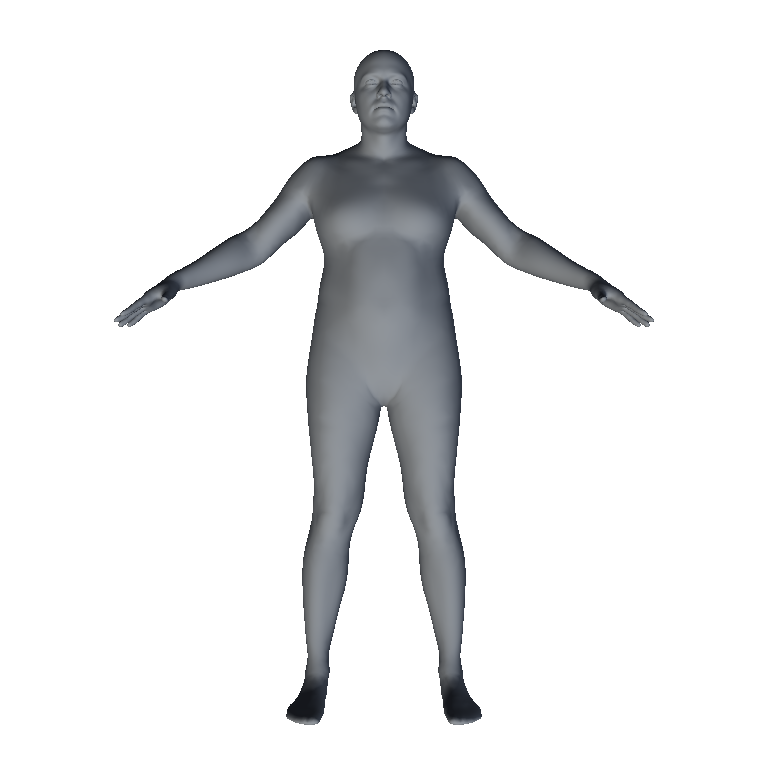}\hfill
    \includegraphics[width=0.49\linewidth]{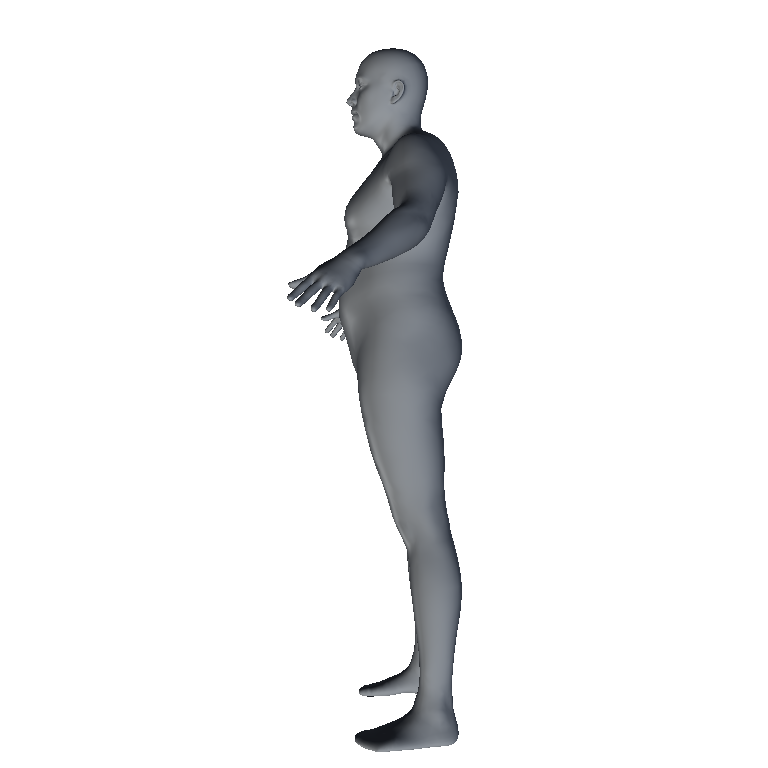}
    \caption{Front and side views of the MHR rest-pose body mesh $\mathcal{M}_0$.}
    \label{fig:rest_pose_mesh}
\end{figure}

Figure~\ref{fig:rest_pose_mesh} visualizes the rest-pose body mesh $\mathcal{M}_0$ of the MHR template used in our avatar representation.

\section{Adaptive UV Resolution}
\label{app:adaptive_uv_resolution}

ARG-Avatar uses connected components of the valid UV domain to adapt the sampling resolution before rasterization.  Specifically, we identify the disconnected UV islands, associate each island with its corresponding anatomical region, and rasterize each component independently at a component-specific scale.  Starting from a base resolution of $512$, we use $1.0\times$ resolution for the face, $0.5\times$ for the hands, and $2.0\times$ for the body, limbs, and feet.  We patchify the components separately, producing $N_T$ UV tokens in total, and apply the CNN decoder independently to each component's feature map; this prevents convolution and patch formation from mixing unrelated UV islands across empty atlas space.  The valid samples from all decoded components form the $L$ dense Gaussian primitives; each sample retains its face index and barycentric coordinates, so the adaptive sampling does not alter its surface anchor.  This component-wise construction increases token density where clothing and articulation induce larger geometric variation, while avoiding unnecessary samples in compact regions.

\section{LBS Animation for UV 3DGS}
\label{app:lbs_details}
\label{app:uv_animation}

Here, we describe in detail how ARG-Avatar uses linear blend skinning (LBS) to animate UV-anchored Gaussian primitives. The goal of the forward LBS operator $\mathrm{LBS}_t$ in~\cref{sec:method:forward_lbs,eq:lbs_operator} is to convert a canonical avatar $\Cm$, represented in the rest pose of the MHR template, into a posed avatar $\Gm_t$ under the target skeletal pose $\theta_t$. Because every Gaussian is anchored to a UV sample on the template surface, the animation can be defined by first animating the underlying mesh and then transferring the resulting rigid motion to each UV sample.

\paragraph{Canonical UV anchoring.}
The MHR rest-pose mesh is $\mathcal{M}_0=(\mathcal{V}_0,\mathcal{F})$, with vertices $\mathcal{V}_0=\{\mathbf{v}_i\}$ and triangular faces $\mathcal{F}$. Each valid UV texel corresponds to one triangle $f_k=(i_1,i_2,i_3)$ and barycentric coordinates $\boldsymbol{\beta}_k=(\beta_{k1},\beta_{k2},\beta_{k3})$ on that triangle. The rest-pose surface anchor of Gaussian $k$ is therefore
\begin{equation}
    \mathbf{a}_k =
    \sum_{r=1}^{3}\beta_{kr}\,\mathbf{v}_{i_r}.
    \label{eq:uv_anchor_appendix}
\end{equation}
ARG-Avatar predicts a canonical offset $\ov_k$ and Gaussian attributes $(\qv_k,\sv_k,\alpha_k,\shv_k)$ at this UV location, so the canonical Gaussian center is $\muv_k=\mathbf{a}_k+\ov_k$. Pose-dependent residuals, when enabled, update the canonical offset, orientation, and scale before the LBS animation is applied, as in~\cref{eq:posed_avatar}.

\paragraph{Skeleton transforms.}
The skeleton $\mathcal{B}$ has joints $\{\mathbf{r}_j\}_{j=1}^{J}$, a kinematic-tree parent map $\pi(j)$, target joint rotations $\theta_t=\{\theta_{t,j}\}_{j=1}^{J}$, and mesh skinning weights $W^m=[w^m_{i,j}]$. For each joint, LBS first constructs a local transform
\begin{equation}
    \mathbf{L}_{t,j} =
    \begin{bmatrix}
        \mathbf{R}(\theta_{t,j}) & \mathbf{b}_j \\
        \mathbf{0}^{\top} & 1
    \end{bmatrix},
    \label{eq:skeleton_local_transform}
\end{equation}
where $\mathbf{R}(\theta_{t,j})$ is the joint rotation, $\mathbf{b}_j=\mathbf{r}_j$ for the root, and $\mathbf{b}_j=\mathbf{r}_j-\mathbf{r}_{\pi(j)}$ otherwise.
The global transformation $\mathbf{T}_{t,j}$ of each joint is obtained by accumulating these local transforms along the kinematic tree, $\mathbf{T}_{t,j}=\mathbf{T}_{t,\pi(j)}\,\mathbf{L}_{t,j}$, starting from the root.
The rest-pose global transformation $\mathbf{T}_{c,j}$ is computed in the same way from the rest-pose parameters.
The transform that moves geometry from the rest pose to pose $t$ is then the relative joint transform $\mathbf{T}_{t,j}\mathbf{T}_{c,j}^{-1}$.

\paragraph{Per-vertex mesh animation.}
For each mesh vertex $i$, standard LBS blends the relative joint transforms with the vertex skinning weights:
\begin{equation}
    \mathbf{T}_{t,i}^{m}
    =
    \sum_{j=1}^{J}
    w^m_{i,j}\,\mathbf{T}_{t,j}\mathbf{T}_{c,j}^{-1}.
    \label{eq:mesh_lbs_appendix}
\end{equation}
Applying this blended transform to the rest vertex gives the posed mesh vertex
\begin{equation}
    \tilde{\mathbf{v}}_{t,i}
    =
    \mathbf{T}_{t,i}^{m}
    \begin{bmatrix}
        \mathbf{v}_i \\
        1
    \end{bmatrix}.
    \label{eq:posed_vertex_appendix}
\end{equation}
We obtain a quaternion estimate from the upper-left $3\times3$ block of each blended per-vertex transform and select its sign using the scalar component $w$:
\begin{equation}
    \begin{aligned}
    \tilde{\qv}_{t,i} &= \operatorname{quat}\!\left((\mathbf{T}_{t,i}^{m})_{1:3,1:3}\right), \\
    \qv_{t,i} &= \operatorname{sign}(\tilde q_{t,i,w})\,\tilde{\qv}_{t,i}.
    \end{aligned}
    \label{eq:vertex_quat_appendix}
\end{equation}
The blended block is generally nonorthogonal. The implementation applies matrix-to-quaternion conversion directly, with no separate projection onto $\mathrm{SO}(3)$. The scalar-sign convention is applied before interpolation, without additional per-face sign matching; unit normalization is performed after interpolation below.

\paragraph{Transfer from mesh vertices to UV Gaussians.}
A UV-anchored Gaussian is not itself a mesh vertex; it lies at a barycentric location on a template triangle. We therefore transfer the mesh animation to each Gaussian by barycentric interpolation over the same face used to define its UV anchor. For Gaussian $k$ on face $f_k=(i_1,i_2,i_3)$, the posed surface anchor is
\begin{equation}
    \tilde{\mathbf{a}}_{t,k}
    =
    \sum_{r=1}^{3}\beta_{kr}\,\tilde{\mathbf{v}}_{t,i_r}.
    \label{eq:posed_uv_anchor_appendix}
\end{equation}
The corresponding rigid orientation is obtained by barycentrically blending the per-vertex quaternions and renormalizing:
\begin{equation}
    \qv^{\mathrm{rigid}}_{t,k}
    =
    \operatorname{norm}
    \left(
        \sum_{r=1}^{3}\beta_{kr}\,\qv_{t,i_r}
    \right).
    \label{eq:uv_rigid_quat_appendix}
\end{equation}
This gives every UV Gaussian a posed surface location and a local rigid rotation induced by the animated template.

\paragraph{Animating Gaussian centers and orientations.}
The canonical offset $\ov_k$ represents geometry that departs from the template surface, such as clothing thickness or hair. We rotate this offset by the interpolated rigid motion and add it to the posed surface anchor:
\begin{equation}
    \mathbf{p}_{t,k}
    =
    \tilde{\mathbf{a}}_{t,k}
    +
    \qv^{\mathrm{rigid}}_{t,k}\cdot\ov_k.
    \label{eq:posed_gaussian_center_appendix}
\end{equation}
The Gaussian orientation is animated by composing the rigid LBS rotation with the canonical Gaussian rotation:
\begin{equation}
    \qv_{t,k}
    =
    \qv^{\mathrm{rigid}}_{t,k}\otimes\qv_k.
    \label{eq:posed_gaussian_quat_appendix}
\end{equation}
After any pose-dependent residuals are applied, the scale, opacity, and spherical-harmonic color coefficients are retained during animation. Thus, the posed avatar is
\begin{equation}
    \Gm_t
    =
    \left\{
        (\mathbf{p}_{t,k},\qv_{t,k},\sv_k,\alpha_k,\shv_k)
    \right\}_{k=1}^{L}.
    \label{eq:posed_gaussian_set_appendix}
\end{equation}
This is the UV-anchored Gaussian version of the LBS operator in~\cref{eq:lbs_operator}. In short, ARG-Avatar keeps the avatar representation dense and image-like in UV space, but uses the template mesh and skeleton to provide a physically meaningful animation field for all Gaussian primitives.

\section{Loss Details}
\label{app:loss_details}

ARG-Avatar is trained with photometric losses to reconstruct the input and driving-pose images, following other avatar reconstruction methods~\cite{zhuang2024idol,qiu2025lhmb,qiu2025lhm}. We introduce the proposed regularizer in the main paper; here, we provide the complete loss formulation. The photometric matching loss is
\begin{equation}
    \mathcal{L}_{\mathrm{pm}}
    = \lambda_1\mathcal{L}_{\mathrm{rgb}}
    + \lambda_{\mathrm{ssim}}\mathcal{L}_{\mathrm{SSIM}}
    + \lambda_{\mathrm{lpips}}\mathcal{L}_{\mathrm{LPIPS}}
    + \lambda_{\mathrm{mask}}\mathcal{L}_{\mathrm{mask}}.
    \label{eq:photo_loss}
\end{equation}
Here, $\mathcal{L}_{\mathrm{rgb}}$ is the SmoothL1 loss with transition parameter $\beta=0.05$, evaluated after compositing the prediction and target with the same background outside the ground-truth foreground mask. The mask term $\mathcal{L}_{\mathrm{mask}}$ also uses SmoothL1 with $\beta=0.05$ to compare the rendered alpha mask with the ground-truth foreground mask.  We additionally apply the photometric loss to a face crop whose bounding box is derived from the avatar's projected face vertices; this term is denoted by $\mathcal{L}_{\mathrm{face}}$.  When ITR is enabled, we supervise the token avatars with the same objective; when pose-dependent deformations are enabled, we also supervise the canonical avatar, obtained by setting every pose-dependent residual to zero:
\begin{equation}
    \mathcal{L}_{\mathrm{cano}}^{m}
    = \frac{1}{V}\sum_{v=1}^{V}
    \left(\mathcal{L}_{\mathrm{pm,\,cano}}^{m,v}
    + \lambda_{\mathrm{face}}\mathcal{L}_{\mathrm{face,\,cano}}^{m,v}\right).
    \label{eq:canonical_animation_loss}
\end{equation}
For dense pass $m$, the rendering loss averages the full-image and face-crop terms over the $V$ supervision views:
\begin{equation}
    \mathcal{L}_{\mathrm{render}}^{m}
    = \frac{1}{V}\sum_{v=1}^{V}
    \left(\mathcal{L}_{\mathrm{pm}}^{m,v}
    + \lambda_{\mathrm{face}}\mathcal{L}_{\mathrm{face}}^{m,v}\right).
    \label{eq:render_loss}
\end{equation}

\paragraph{Intermediate token-rendering supervision.}
With intermediate token rendering (ITR), each intermediate token representation is decoded into a token-GS avatar, animated, and rendered in every supervision view.  Its reconstruction loss enters the objective as an additional $\mathcal{L}_{\mathrm{render}}^{m}$.  For these coarse token-GS passes, we omit LPIPS and face-crop supervision.  All rendered passes are averaged uniformly, so intermediate predictions receive the same aggregate weight as the final dense prediction.

\paragraph{Normal Offset Regularization.}
We encourage each canonical offset $\ov_i$, sampled from the UV map, to align with its corresponding template-surface normal axis $\nv_i$. The penalty treats parallel and antiparallel offsets equally. To avoid numerical instability, we evaluate this term only for offsets satisfying $\|\ov_i\|_2 > 10^{-3}$:
\begin{equation}
    \mathcal{L}_{\text{normal}}
    = \frac{1}{|\mathcal{V}_n|}
    \sum_{i\in\mathcal{V}_n} \left(1 - \cos(\ov_i, \nv_i)^2\right),
    \label{eq:normal_loss}
\end{equation}
where $\mathcal{V}_n=\{i:\|\ov_i\|_2>10^{-3}\}$.

\paragraph{UV Deformation Smoothness.}
We encourage spatially smooth Gaussian attributes in UV space.  The regularized channels include the canonical offsets $\ov$, scales $\sv$, rotations $\qv$, pose-dependent deformations $\dv_v=[\delta\ov_v,\delta\sv_v,\delta\qv_v]$, and the corresponding offset magnitudes.  We concatenate them as
\begin{equation}
    \fv = [\ov, \sv, \qv, \{\dv_v\}_{v=1}^{V}, \|\ov\|_2, \{\|\delta\ov_v\|_2\}_{v=1}^{V}],
\end{equation}
and apply a discrete UV Laplacian $\Delta_{\text{uv}}$ over valid UV texels:
\begin{equation}
    \mathcal{L}_{\text{smooth}}
    = \frac{1}{|\Omega_{\text{valid}}|}
    \|\Delta_{\text{uv}}\fv\|_F^2.
    \label{eq:smooth_loss}
\end{equation}
The Laplacian is evaluated only at texels with a valid $3\times3$ UV neighborhood.  For the connected-component UV representation, we apply it independently within each component.  Before applying the Laplacian, we scale the offset-magnitude channels by $\lambda_{\mathrm{offset\text{-}norm}}$.

\paragraph{Offset Magnitude.}
We penalize large canonical offsets with an L2 term:
\begin{equation}
    \mathcal{L}_{\text{L2}}
    = \frac{1}{L}\sum_i \|\ov_i\|_2^2.
    \label{eq:l2_loss}
\end{equation}

\paragraph{Pose-deformation regularization.}
We penalize the pose-dependent position offsets, raw scale residuals, and deviations from the identity rotation:
\begin{equation}
    \mathcal{L}_{\mathrm{pose}}
    = \frac{1}{VL}\sum_{v,i}\left(
    \|\delta\ov_{v,i}\|_2^2
    + \|\mathbf{r}^{s}_{v,i}\|_2^2
    + \|\delta\qv_{v,i}-\mathbf{1}_q\|_2^2\right),
    \label{eq:pose_reg_loss}
\end{equation}
where $\mathbf{1}_q$ is the identity quaternion and $\mathbf{r}^{s}_{v,i}\in\mathbb{R}^{3}$ is the raw scale residual, related to the multiplicative factor in Eq.~\ref{eq:posed_avatar} by $\delta\sv_{v,i}=\mathbf{1}+\tanh(\mathbf{r}^{s}_{v,i})$, with $\tanh$ applied elementwise.  We further apply the common shell-transfer loss $\mathcal{L}_{\mathrm{shell}}$ defined in~\cref{eq:shell_loss}.  It estimates the component shared by the per-view pose offsets, transfers this component to the canonical offset with stop-gradient targets, and removes it from the pose offsets.  The loss therefore assigns view-shared geometry to the canonical avatar instead of representing it redundantly through pose residuals.

\paragraph{Token and pixel residual regularization.}
The token branch predicts a 3D offset $\boldsymbol{\tau}_{v,k}$ for each UV token $k$. We broadcast it to the dense samples in that token's patch, accumulate the sample offsets onto mesh vertices using normalized barycentric weights, and interpolate the resulting vertex deformation back to the Gaussian anchors. Let $\kappa(i)$ be the token containing dense sample $i$, and let $b_{ij}$ be its barycentric weight for mesh vertex $j$, with $b_{ij}=0$ outside its anchor triangle. The vertex-lifted token deformation and final pose offset are
\[
    \begin{aligned}
    \delta\mathbf{v}_{v,j}
    &= \frac{\sum_{i=1}^{L}b_{ij}\boldsymbol{\tau}_{v,\kappa(i)}}
    {\sum_{i=1}^{L}b_{ij}+10^{-8}}, \\
    \delta\ov_{v,i}
    &= \sum_{j\in f_i}b_{ij}\,\delta\mathbf{v}_{v,j}
    + \boldsymbol{\epsilon}_{v,i},
    \end{aligned}
\]
where $\boldsymbol{\epsilon}_{v,i}$ is the pixel-level residual predicted by the dense pose decoder $\mathrm{PDec}$. The canonical offset uses the same token-to-vertex-to-sample lifting with a separate dense component from $\mathrm{CDec}$.
For the vertex-lifted token deformation $\delta\mathbf{v}_{v,i}$, we penalize changes in the relative lengths of template-mesh edges $(i,j)$:
\begin{equation}
    \begin{aligned}
    \mathcal{L}_{\mathrm{strain}}
    ={}& \frac{1}{V|\mathcal{E}|}\sum_{v,(i,j)\in\mathcal{E}}
    \left(
    \frac{
    \|\mathbf{v}_i+\delta\mathbf{v}_{v,i}
    -\mathbf{v}_j-\delta\mathbf{v}_{v,j}\|_2}
    {\|\mathbf{v}_i-\mathbf{v}_j\|_2}
    \right. \\
    &\left.\hspace{4.8em}
    - 1
    \right)^2.
    \end{aligned}
    \label{eq:token_strain_loss}
\end{equation}
For pixel-level residuals $\boldsymbol{\epsilon}_{v,i}$ and their canonical counterparts, we aggregate the residuals within each UV patch and penalize the squared norm of the patch mean, yielding $\mathcal{L}_{\mathrm{pixel}}$.  This separates the roles of the two branches: the token branch models patch-level deformation, while the pixel branch is reserved for local detail.

\paragraph{Scale Range.}
For each predicted Gaussian scale value $s$, we penalize values outside a stable range and softly discourage overly large positive scales:
\begin{equation}
    \ell_{\text{scale}}(s)
    = \max\left(
        f_+(s-3) + f_+(-2-s),
        0.1\,f_+(s)
    \right),
    \label{eq:scale_penalty}
\end{equation}
where $f_+(x)=\max(x,0)$.  We define $\mathcal{L}_{\text{scale}}$ as the mean of $\ell_{\text{scale}}$ over all scale channels.

\paragraph{Full objective.}
The regularization term combines the preceding attribute constraints:
\begin{equation}
    \begin{aligned}
    \mathcal{L}_{\text{reg}}
    ={}& \lambda_n\,\mathcal{L}_{\text{normal}}
    + \lambda_s\,\mathcal{L}_{\text{smooth}}
    + \lambda_{\text{l2}}\,\mathcal{L}_{\text{L2}}
    + \lambda_{\text{pose}}\,\mathcal{L}_{\text{pose}} \\
    &+ \lambda_{\text{shell}}\,\mathcal{L}_{\text{shell}}
    + \lambda_{\text{strain}}\,\mathcal{L}_{\text{strain}}
    + \lambda_{\text{pixel}}\,\mathcal{L}_{\text{pixel}}
    + \lambda_{\text{sc}}\,\mathcal{L}_{\text{scale}}.
    \end{aligned}
    \label{eq:reg_loss}
\end{equation}
Let $M$ be the number of rendered passes, including the final dense pass and, when used, ITR passes.  The complete training objective is
\begin{equation}
    \mathcal{L}
    = \frac{1}{M}\sum_{m=1}^{M}
    \left(
        \mathcal{L}_{\mathrm{render}}^{m}
        + \lambda_{\mathrm{cano}}\mathcal{L}_{\mathrm{cano}}^{m}
        + \mathcal{L}_{\mathrm{reg}}^{m}
    \right).
    \label{eq:full_loss}
\end{equation}
Here, $\mathcal{L}_{\mathrm{cano}}^{m}=0$ for passes without pose-dependent deformations, and $\mathcal{L}_{\mathrm{reg}}^{m}=0$ for non-dense passes.  Thus, regularization is applied only to dense UV-GS predictions.

\paragraph{Coefficients.}
Across all experiments, we set $\lambda_1=0.8$, $\lambda_{\text{ssim}}=0.2$, $\lambda_{\text{lpips}}=1.0$, $\lambda_{\text{mask}}=0.5$, $\lambda_{\text{face}}=1.0$, and $\lambda_{\text{cano}}=0.5$.  ITR has no separate multiplier because its render passes are included in the uniform average in~\cref{eq:full_loss}.  For regularization, we use $\lambda_n=0.1$, $\lambda_s=10^{-4}$, $\lambda_{\mathrm{offset\text{-}norm}}=0.1$, $\lambda_{\text{l2}}=0.1$, $\lambda_{\text{pose}}=1.0$, $\lambda_{\text{shell}}=10^{-3}$, $\lambda_{\text{strain}}=1.0$, $\lambda_{\text{pixel}}=1.0$, and $\lambda_{\text{sc}}=1.0$.  We selected these coefficients in preliminary trials and fixed them for every experiment.

\section{Training Details}
\label{app:training_details}

Each training batch samples six input observations and six target supervision views from the same subject. The model is trained for 100K steps with AdamW, cosine learning-rate scheduling, and a global batch size of 96. Training uses bf16 mixed precision. The SAM3DBody backbone is frozen. The final model is trained on four RTX A6000 GPUs for approximately 200 hours, corresponding to 125 H100-equivalent GPU-hours under the conversion used in~\cref{app:scale_training_compute}. For the ablation study, every variant uses $512^2$ UV resolution without adaptive UV resolution and is trained for 50K steps.

\paragraph{Gaussian Decoders.}
Both $\mathrm{CDec}$ and $\mathrm{PDec}$ take 1024-dimensional tokens arranged in UV space. A linear layer and a $1\times1$ convolution project the token map to 256 channels, followed by three residual blocks with 256, 128, and 64 output channels. Each block uses GroupNorm and SiLU and is followed by $2\times$ upsampling and a convolution, giving $8\times$ upsampling in total. A final GroupNorm, SiLU, and $1\times1$ convolution predict the Gaussian attributes or pose-dependent residuals. The two decoders have separate weights and operate independently on each UV component.

\paragraph{Training Dataset.}
Our training set is composed entirely of public human datasets and contains 6,821 subjects and 64,142 sequences in total.
Each sequence corresponds to either a subject in a 3D scan dataset or a clip in a monocular or multiview video dataset.
DNARendering~\cite{cheng2023dnarendering} contributes 313 subjects and 53,100 motion and camera clips, providing the largest source of calibrated multiview dynamic supervision. We use only Parts 1--4 because of storage limitations. 4D-Dress~\cite{wang2024ddress} adds 27 subjects and 972 sequences with challenging garment deformation. HumanVid~\cite{wang2024humanvid} provides 2,537 human video clips. We further include scan data from 2K2K~\cite{han2023highfidelity} and THuman2.1~\cite{yu2021function}, which contain 1,732 and 2,135 subjects, respectively, and generate synthetic views from these scans.
Human4DiT~\cite{shao2024human} consists of 77 subjects in different outfits, totaling 3,666 scans. For every training dataset, we reserve 10\% of subjects as a held-out validation split and exclude them from training. NeuMan~\cite{jiang2022neuman} is never used in training or validation.

\section{Scale and Training Compute}
\label{app:scale_training_compute}

\begin{table}[t]
    \centering
    \caption{Training dataset size comparisons. ``N/A'' refers to information not publicly available.}
    \vspace{-5pt}
    \label{tab:dataset_statistics}
    \small
    \setlength{\tabcolsep}{3pt}
    \renewcommand{\arraystretch}{1.12}
    \resizebox{\columnwidth}{!}{%
    \begin{tabular}{l >{\raggedright\arraybackslash}p{5cm} l}
        \toprule
        Method & Training datasets & Seqs. \\
        \midrule
        IDOL~\cite{zhuang2024idol} & HuGe100K; THuman 2.1 & 102K \\
        \midrule
        LHM~\cite{qiu2025lhm} & \multirow{2}{=}{\raggedright Private data; 2K2K, Human4DiT, RenderPeople} & \multirow{2}{*}{307K} \\
        LHM++~\cite{qiu2025lhmb} & & \\
        \midrule
        ARG-Avatar & \multirow{2}{=}{\raggedright DNARendering, HumanVid, 2K2K, 4D-Dress, Human4DiT, THuman2.1} & \multirow{2}{*}{64K} \\
        (Ours) & & \\
        \bottomrule
    \end{tabular}
    }
    \vspace{-10pt}
\end{table}

\begin{table}[t]
    \centering
    \caption{Comparison of model size and training compute for models evaluated on NeuMan.}
    \label{tab:efficiency}
    \small
    \setlength{\tabcolsep}{2pt}
    \renewcommand{\arraystretch}{1.15}
    \resizebox{\columnwidth}{!}{%
    \begin{tabular}{lccccc}
        \toprule
        \multirow[c]{2}{*}{Method} & Trainable & Backbone & Total & H100 GPU & Training \\
         & Params & Params & Params & hours & Hardware \\
        \midrule
        IDOL~\cite{zhuang2024idol} & 544M & 1.2B & 1.7B & 768 & 32$\times$H100 \\
        LHM~\cite{qiu2025lhm} & 1.1B & 1.5B & 2.6B & 3814 & 64$\times$A100 \\
        LHM++~\cite{qiu2025lhmb} & 609M & 330M & 939M & 1695 & 32$\times$A100 \\
        Ours & 68M & 841M & 908M & 125 & 4$\times$A6000 \\
        \bottomrule
    \end{tabular}
    }
\end{table}

Table~\ref{tab:dataset_statistics} reports the training-data scale of ARG-Avatar and the compared feedforward baselines.
Table~\ref{tab:efficiency} compares model size and training compute. ARG-Avatar is trained for 125 H100-equivalent GPU-hours, which is 7\% of LHM++'s training-compute budget. We convert GPU-hours across devices to approximate H100-equivalent GPU-hours using the official bf16 compute capabilities: 1,979 TFLOPS, 624 TFLOPS, and 310 TFLOPS for H100, A100, and A6000, respectively.

\section{Runtime Profile}
\label{app:runtime_profile}

\begin{table}[t]
    \centering
    \caption{NeuMan runtime profile for IDOL, LHM, LHM++, and ARG-Avatar. Reconstruction time measures feedforward avatar reconstruction latency; animation FPS measures rendering throughput over 104 frames; peak VRAM is the peak GPU memory usage during the profiled run.}
    \label{tab:profile}
    {\scriptsize
    \setlength{\tabcolsep}{2pt}
    \begin{tabular}{>{\raggedright\arraybackslash}p{1.05cm}cccc}
        \whline{1.2pt}
        \multirow[c]{2}{*}{Method} & \multirow[c]{2}{*}{$V$} & Reconstruction & Animation & GPU \\
         & & (s) $\downarrow$ & FPS $\uparrow$ & VRAM $\downarrow$ \\
        \hline
        \parbox[c]{1.05cm}{\raggedright IDOL \\ \cite{zhuang2024idol}} & 1 & 2.5 & 45 & 17.0 \\
        \whline{0.6pt}
        \noalign{\vskip 3pt}
        \parbox[c]{1.05cm}{\raggedright LHM~\cite{qiu2025lhm}} & 1 & 20.6 & 85 & 22.0 \\
        \whline{0.6pt}
        \noalign{\vskip 3pt}
        \multirow[c]{5}{*}{\parbox[c]{1.05cm}{\raggedright LHM++ \\ \cite{qiu2025lhmb}}} & 1 & 2.8 & \multirow[c]{5}{*}{20} & 7.9 \\
         & 2 & 4.5 &  & 7.9 \\
         & 3 & 6.6 &  & 8.1 \\
         & 4 & 8.8 &  & 8.4 \\
         & 6 & 14.2 &  & 9.1 \\
        \whline{0.6pt}
        \noalign{\vskip 3pt}
        \multirow[c]{5}{*}{\parbox[c]{1.05cm}{\raggedright Ours}} & 1 & 1.8 & \multirow[c]{5}{*}{80} & 7.7 \\
         & 2 & 1.8 &  & 7.7 \\
         & 3 & 1.9 &  & 8.0 \\
         & 4 & 1.9 &  & 8.4 \\
         & 6 & 2.3 &  & 9.1 \\
        \whline{1.2pt}
    \end{tabular}
    }
\end{table}

\begin{figure}[H]
    \centering
    \includegraphics[width=0.62\linewidth]{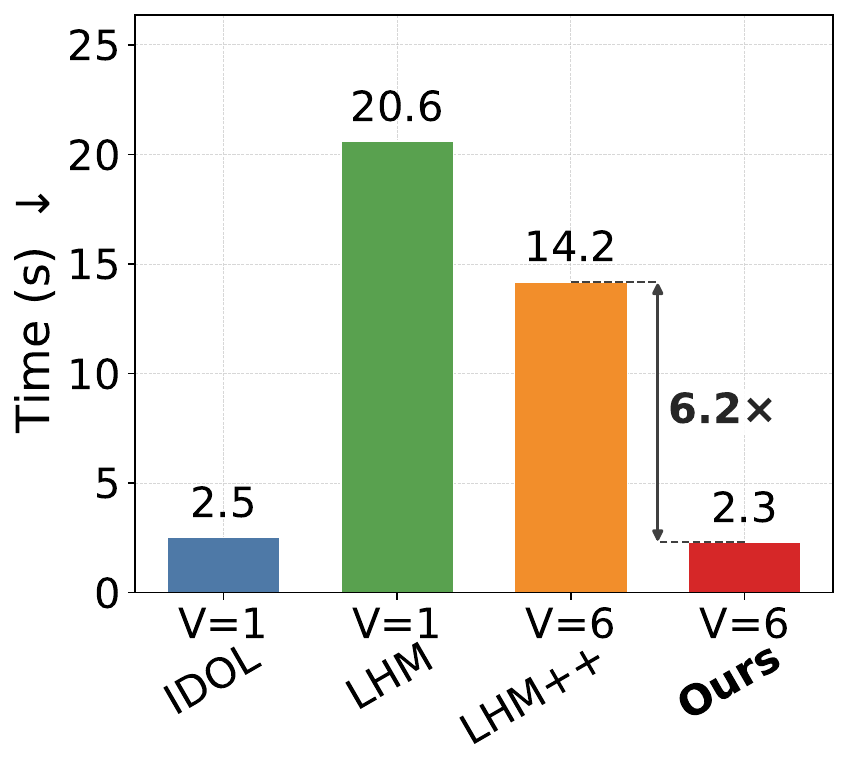}
    \caption{Feedforward reconstruction latency on NeuMan for the $V=1$ IDOL and LHM baselines and the $V=6$ LHM++ and ARG-Avatar configurations. At six views, ARG-Avatar reconstructs an avatar $6.2\times$ faster than LHM++.}
    \label{fig:runtime_recon_time}
\end{figure}

\begin{table*}[t]
    \centering
    \caption{Per-dataset results on held-out validation splits for novel-pose or novel-view synthesis. Vertical rules separate datasets. ``Overall'' aggregates results across all evaluated datasets, including NeuMan (reported in the main paper), weighted by the number of cross-samples. \protect\colorbox[rgb]{0.957,0.490,0.459}{First}, \protect\colorbox[rgb]{1.000,0.922,0.400}{second}, and \protect\colorbox[rgb]{1.000,0.949,0.800}{third} indicate the top three results in each metric column.}
    \label{tab:full_cross_metrics}
    {\tiny
    \setlength{\tabcolsep}{2pt}
    \setlength{\arrayrulewidth}{0.3pt}
    \resizebox{\textwidth}{!}{%
    \begin{tabular}{ll|ccc|ccc|ccc|ccc|ccc|ccc|ccc}
        \whline{1.2pt}
        Method & Views & \multicolumn{3}{c|}{\textit{2K2K}} & \multicolumn{3}{c|}{\textit{THuman2.1}} & \multicolumn{3}{c|}{\textit{Human4DiT}} & \multicolumn{3}{c|}{\textit{4D-Dress}} & \multicolumn{3}{c|}{\textit{DNARendering}} & \multicolumn{3}{c|}{\textit{HumanVid}} & \multicolumn{3}{c}{\textit{Overall}} \\
        \cline{3-23}
         & & PSNR $\uparrow$ & SSIM $\uparrow$ & LPIPS $\downarrow$ & PSNR & SSIM & LPIPS & PSNR & SSIM & LPIPS & PSNR & SSIM & LPIPS & PSNR & SSIM & LPIPS & PSNR & SSIM & LPIPS & PSNR & SSIM & LPIPS \\
        \hline
        SIGMAN & 1 & 15.14 & 0.905 & 0.111 & 15.33 & 0.910 & 0.111 & 15.52 & 0.912 & 0.129 & 14.73 & 0.897 & 0.154 & 16.68 & 0.947 & 0.070 & 16.17 & 0.862 & 0.162 & 15.90 & 0.897 & 0.125 \\
        \hline
        IDOL & 1 & 20.88 & 0.928 & 0.085 & 21.05 & 0.935 & 0.095 & 22.15 & 0.942 & 0.077 & 20.63 & 0.929 & 0.092 & 23.44 & 0.964 & 0.048 & 20.53 & 0.899 & 0.129 & 21.25 & 0.924 & 0.098 \\
        \hline
        LHM & 1 & 22.55 & 0.935 & 0.072 & 22.81 & 0.940 & 0.087 & 23.17 & \cellcolor[rgb]{0.957,0.490,0.459}0.946 & 0.071 & 21.65 & 0.934 & 0.082 & 25.75 & 0.968 & 0.037 & 21.23 & 0.906 & 0.111 & 22.38 & 0.931 & 0.085 \\
        \hline
        \multirow[c]{5}{*}{LHM++} & 1 & 22.24 & 0.935 & 0.070 & 22.34 & 0.939 & 0.086 & 22.36 & 0.943 & 0.074 & 21.55 & 0.934 & 0.079 & 25.80 & 0.968 & 0.036 & 22.03 & 0.910 & 0.104 & 22.62 & 0.932 & 0.082 \\
         & 2 & 22.23 & 0.936 & 0.068 & 22.50 & 0.939 & 0.084 & 22.52 & \cellcolor[rgb]{1.000,0.949,0.800}0.944 & 0.072 & 21.53 & 0.934 & 0.078 & 25.78 & 0.968 & 0.035 & 22.35 & 0.911 & 0.101 & 22.82 & 0.932 & 0.079 \\
         & 3 & 22.33 & 0.936 & 0.067 & 22.59 & 0.940 & 0.083 & 22.66 & \cellcolor[rgb]{1.000,0.949,0.800}0.944 & 0.071 & 21.62 & 0.934 & 0.077 & 25.99 & \cellcolor[rgb]{1.000,0.949,0.800}0.969 & 0.034 & 22.40 & 0.911 & 0.100 & 22.91 & 0.933 & 0.078 \\
         & 4 & 22.38 & 0.936 & 0.066 & 22.58 & 0.940 & 0.082 & 22.79 & \cellcolor[rgb]{1.000,0.949,0.800}0.944 & 0.070 & 21.63 & 0.934 & 0.077 & 26.10 & \cellcolor[rgb]{1.000,0.949,0.800}0.969 & 0.034 & 22.47 & 0.911 & 0.099 & 22.98 & 0.933 & 0.077 \\
         & 6 & 22.21 & 0.936 & 0.068 & 22.41 & 0.940 & 0.083 & 22.87 & \cellcolor[rgb]{1.000,0.922,0.400}0.945 & 0.070 & 21.69 & 0.934 & 0.077 & 26.18 & \cellcolor[rgb]{1.000,0.949,0.800}0.969 & 0.033 & 22.50 & 0.912 & 0.098 & 22.98 & 0.933 & 0.077 \\
        \hline
        \multirow[c]{5}{*}{Ours} & 1 & 23.58 & 0.932 & 0.057 & 24.46 & 0.939 & 0.061 & 24.21 & 0.941 & 0.055 & 23.59 & 0.934 & 0.056 & 26.25 & 0.965 & 0.032 & 21.64 & 0.897 & 0.100 & 23.13 & 0.925 & 0.071 \\
         & 2 & 25.29 & 0.938 & 0.043 & 26.20 & 0.943 & 0.048 & 25.34 & 0.942 & 0.046 & 24.52 & \cellcolor[rgb]{1.000,0.949,0.800}0.936 & 0.048 & 28.11 & 0.968 & 0.025 & 24.57 & 0.912 & 0.076 & 25.45 & 0.934 & 0.055 \\
         & 3 & \cellcolor[rgb]{1.000,0.949,0.800}26.13 & \cellcolor[rgb]{1.000,0.949,0.800}0.940 & \cellcolor[rgb]{1.000,0.949,0.800}0.038 & \cellcolor[rgb]{1.000,0.949,0.800}26.86 & \cellcolor[rgb]{1.000,0.949,0.800}0.945 & \cellcolor[rgb]{1.000,0.949,0.800}0.044 & \cellcolor[rgb]{1.000,0.949,0.800}25.75 & 0.943 & \cellcolor[rgb]{1.000,0.949,0.800}0.044 & \cellcolor[rgb]{1.000,0.949,0.800}24.96 & \cellcolor[rgb]{1.000,0.922,0.400}0.937 & \cellcolor[rgb]{1.000,0.949,0.800}0.045 & \cellcolor[rgb]{1.000,0.949,0.800}28.81 & \cellcolor[rgb]{1.000,0.922,0.400}0.970 & \cellcolor[rgb]{1.000,0.949,0.800}0.023 & \cellcolor[rgb]{1.000,0.949,0.800}25.03 & \cellcolor[rgb]{1.000,0.949,0.800}0.914 & \cellcolor[rgb]{1.000,0.949,0.800}0.072 & \cellcolor[rgb]{1.000,0.949,0.800}25.99 & \cellcolor[rgb]{1.000,0.949,0.800}0.936 & \cellcolor[rgb]{1.000,0.949,0.800}0.051 \\
         & 4 & \cellcolor[rgb]{1.000,0.922,0.400}26.54 & \cellcolor[rgb]{1.000,0.922,0.400}0.941 & \cellcolor[rgb]{1.000,0.922,0.400}0.036 & \cellcolor[rgb]{1.000,0.922,0.400}27.25 & \cellcolor[rgb]{1.000,0.922,0.400}0.947 & \cellcolor[rgb]{1.000,0.922,0.400}0.041 & \cellcolor[rgb]{1.000,0.922,0.400}25.98 & \cellcolor[rgb]{1.000,0.949,0.800}0.944 & \cellcolor[rgb]{1.000,0.922,0.400}0.042 & \cellcolor[rgb]{1.000,0.922,0.400}25.11 & \cellcolor[rgb]{1.000,0.922,0.400}0.937 & \cellcolor[rgb]{1.000,0.922,0.400}0.044 & \cellcolor[rgb]{1.000,0.922,0.400}29.23 & \cellcolor[rgb]{1.000,0.922,0.400}0.970 & \cellcolor[rgb]{1.000,0.922,0.400}0.021 & \cellcolor[rgb]{1.000,0.922,0.400}25.44 & \cellcolor[rgb]{1.000,0.922,0.400}0.916 & \cellcolor[rgb]{1.000,0.922,0.400}0.069 & \cellcolor[rgb]{1.000,0.922,0.400}26.36 & \cellcolor[rgb]{1.000,0.922,0.400}0.937 & \cellcolor[rgb]{1.000,0.922,0.400}0.049 \\
         & 6 & \cellcolor[rgb]{0.957,0.490,0.459}26.75 & \cellcolor[rgb]{0.957,0.490,0.459}0.943 & \cellcolor[rgb]{0.957,0.490,0.459}0.035 & \cellcolor[rgb]{0.957,0.490,0.459}27.43 & \cellcolor[rgb]{0.957,0.490,0.459}0.948 & \cellcolor[rgb]{0.957,0.490,0.459}0.039 & \cellcolor[rgb]{0.957,0.490,0.459}26.30 & \cellcolor[rgb]{1.000,0.922,0.400}0.945 & \cellcolor[rgb]{0.957,0.490,0.459}0.041 & \cellcolor[rgb]{0.957,0.490,0.459}25.44 & \cellcolor[rgb]{0.957,0.490,0.459}0.938 & \cellcolor[rgb]{0.957,0.490,0.459}0.043 & \cellcolor[rgb]{0.957,0.490,0.459}29.52 & \cellcolor[rgb]{0.957,0.490,0.459}0.971 & \cellcolor[rgb]{0.957,0.490,0.459}0.020 & \cellcolor[rgb]{0.957,0.490,0.459}25.86 & \cellcolor[rgb]{0.957,0.490,0.459}0.918 & \cellcolor[rgb]{0.957,0.490,0.459}0.067 & \cellcolor[rgb]{0.957,0.490,0.459}26.69 & \cellcolor[rgb]{0.957,0.490,0.459}0.938 & \cellcolor[rgb]{0.957,0.490,0.459}0.047 \\
        \whline{1.2pt}
    \end{tabular}
    }
    }
\end{table*}

Table~\ref{tab:profile} reports the full NeuMan runtime profile for IDOL, LHM, LHM++, and ARG-Avatar across input-view counts $V$, covering feedforward reconstruction latency, animation FPS, and peak GPU VRAM. ARG-Avatar's reconstruction time stays nearly flat as $V$ grows (from $1.8$~s at $V=1$ to $2.3$~s at $V=6$), whereas LHM++ scales from $2.8$~s to $14.2$~s. ARG-Avatar's animation throughput ($80$ FPS) is $4\times$ that of LHM++ ($20$ FPS) at comparable peak VRAM.
\paragraph{Profiling Setup.}
All methods are profiled on a single NVIDIA RTX A6000 GPU with one subject per batch. We process two NeuMan subjects and discard the first as GPU warm-up, leaving one measured subject per configuration. IDOL and LHM use one input view; LHM++ and ARG-Avatar use $V\in\{1,2,3,4,6\}$. Models are loaded before timing, and CUDA is synchronized before and after each timed stage. We measure wall-clock latency with pose estimates precomputed; model loading, data loading, offline pose estimation, and external input preprocessing are excluded.

\paragraph{Timing Scope.}
For ARG-Avatar, reconstruction timing includes the frozen SAM3DBody backbone, the ConvNeXt image encoder, avatar-token processing, Gaussian decoding, and pose-dependent deformation prediction. Intermediate animation and rendering used by ITR are also included in this stage. Final avatar animation and rendering are timed separately from reconstruction, and image-quality metric computation is excluded from both. The saved profiles report the combined reconstruction latency; backbone feature extraction and the remaining avatar-network forward pass are not timed separately.

\paragraph{Animation and Memory.}
Animation FPS is computed as 104 target frames divided by the measured animation/rendering time. Peak VRAM is the maximum allocated GPU memory during the profiled sample, measured in GiB, with peak-memory statistics reset before each sample.

\section{Per-Dataset Quantitative Results}
\label{app:full_results}

\begin{figure*}[h]
    \centering
    \resizebox{\textwidth}{!}{\hspace*{-2cm}\input{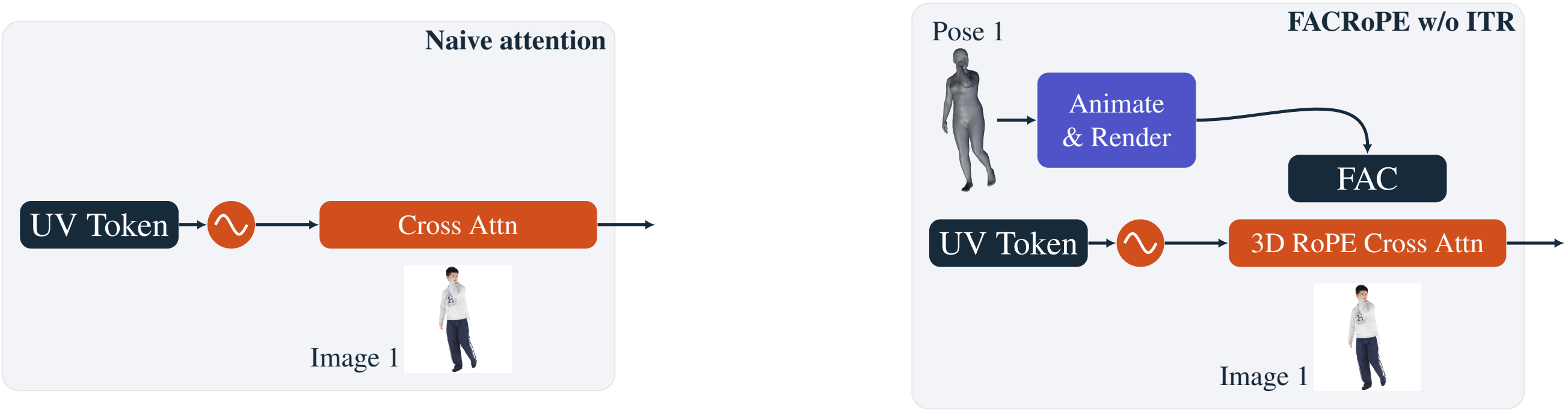}}
    \caption{Ablated ARGBlock variants, drawn with the same visual vocabulary as the full block in~\cref{fig:mvargblock}. \textbf{Naive attention} keeps only the embedding and cross-attention path. \textbf{FACRoPE w/o ITR} animates and renders the template rather than a decoded avatar, so FAC guides the cross-attention but no intermediate avatar is decoded.}
    \label{fig:argblock_variants}
\end{figure*}

\cref{tab:full_cross_metrics} reports the full per-dataset PSNR, SSIM, and LPIPS breakdown for all methods and input-view counts. The detailed table matches the aggregate trend in~\cref{tab:main_results}: ARG-Avatar consistently benefits from additional input observations, improving from 23.13 dB to 26.69 dB overall as $V$ increases from one to six. LHM++ changes much less over the same range, from 22.62 dB to 22.98 dB overall, and is nearly flat on several held-out training datasets. ARG-Avatar achieves the highest PSNR and lowest LPIPS at $V=6$ on every listed dataset, including the out-of-distribution NeuMan benchmark and the held-out 2K2K, THuman2.1, Human4DiT, 4D-Dress, DNARendering, and HumanVid splits.

\section{SAM3DBody Feature Ablation}
\label{app:sam3dbody_ablation}

\begin{table*}[t]
    \centering
    \caption{Ablation of SAM3DBody and DINOv2 image features.}
    \label{tab:main_results_nosam3db}
    {\scriptsize
    \setlength{\tabcolsep}{2pt}
    \setlength{\arrayrulewidth}{0.3pt}
    \renewcommand{\arraystretch}{1.0}
    \resizebox{\textwidth}{!}{%
    \begin{tabular}{>{\raggedright\arraybackslash}p{1.7cm}l|cccccc|ccc|ccc|ccc}
        \whline{1.2pt}
        & & \multicolumn{6}{c|}{\textit{NeuMan}} & \multicolumn{3}{c|}{\textit{Multiview Videos}} & \multicolumn{3}{c|}{\textit{3D Scans}} & \multicolumn{3}{c}{\textit{HumanVid}} \\
        Method & $V$ & \multicolumn{3}{c}{Novel Poses} & \multicolumn{3}{c|}{Input Poses} & \multicolumn{3}{c|}{Novel Poses} & \multicolumn{3}{c|}{Novel Views} & \multicolumn{3}{c}{Novel Poses} \\
         &  & PSNR $\uparrow$ & SSIM $\uparrow$ & LPIPS $\downarrow$ & PSNR & SSIM & LPIPS & PSNR & SSIM & LPIPS & PSNR & SSIM & LPIPS & PSNR & SSIM & LPIPS \\
        \hline
        \multirow[c]{4}{*}{\parbox[c]{1.7cm}{\raggedright ARG-Avatar\\\mbox{w/ SAM3DBody}}} & 1 & 27.02 & 0.966 & 0.024 & 28.67 & 0.974 & 0.016 & 25.76 & 0.963 & 0.034 & 24.02 & 0.939 & 0.059 & 21.64 & 0.897 & 0.100 \\
         & 2 & 29.17 & 0.971 & 0.019 & \textbf{33.88} & \textbf{0.981} & \textbf{0.010} & 27.35 & 0.966 & 0.028 & 25.34 & 0.943 & 0.048 & 24.57 & 0.912 & 0.076 \\
         & 4 & 30.14 & 0.973 & 0.017 & 33.26 & \textbf{0.981} & 0.011 & 28.30 & 0.969 & 0.024 & 26.28 & 0.946 & 0.042 & 25.44 & 0.916 & 0.069 \\
         & 6 & 30.82 & \textbf{0.975} & \textbf{0.016} & 33.30 & \textbf{0.981} & 0.011 & 28.59 & \textbf{0.970} & \textbf{0.023} & 26.45 & 0.947 & \textbf{0.041} & \textbf{25.86} & \textbf{0.918} & \textbf{0.067} \\
        \whline{0.6pt}
        \multirow[c]{4}{*}{\parbox[c]{1.7cm}{\raggedright ARG-Avatar\\\mbox{w/ DINOv2}}} & 1 & 26.69 & 0.967 & 0.024 & 28.21 & 0.973 & 0.019 & 25.88 & 0.964 & 0.033 & 24.35 & 0.940 & 0.056 & 21.50 & 0.900 & 0.101 \\
         & 2 & 28.62 & 0.971 & 0.020 & 33.08 & 0.980 & 0.012 & 27.87 & 0.967 & 0.027 & 26.04 & 0.944 & 0.046 & 24.03 & 0.914 & 0.079 \\
         & 4 & 30.01 & 0.973 & 0.017 & 32.91 & 0.980 & 0.012 & 28.78 & 0.969 & 0.024 & 26.81 & 0.947 & 0.042 & 24.58 & 0.917 & 0.075 \\
         & 6 & \textbf{30.89} & \textbf{0.975} & \textbf{0.016} & 33.11 & 0.980 & 0.012 & \textbf{28.99} & \textbf{0.970} & \textbf{0.023} & \textbf{26.98} & \textbf{0.948} & \textbf{0.041} & 24.85 & \textbf{0.918} & 0.073 \\
        \whline{1.2pt}
    \end{tabular}
    }
    }
\end{table*}

We study the effect of using SAM3DBody features in ARG-Avatar.
Since the baselines use DINOv2~\cite{oquab2023dinov} image features, we replace the SAM3DBody features with DINOv2-Large features in ARG-Avatar.
The model is trained for 100K steps with the same configuration as the main architecture.
\cref{tab:main_results_nosam3db} reports the saved SAM3DBody evaluations and the 47K-step DINOv2 evaluations with $V\in\{1,2,4,6\}$ input views. For SAM3DBody, we use the HumanVid results from the main paper, the released NeuMan results for all reported input-view counts, and the existing 44K-step checkpoint results for the remaining datasets.

\paragraph{Results.}
The DINOv2 variant achieves 30.89 dB on NeuMan with six input views, showing that strong multiview reconstruction can be retained without the SAM3DBody feature branch.
We primarily use SAM3DBody features to avoid running a separate image feature extractor, since SAM3DBody is already run for pose estimation.

\section{Additional Qualitative Results}
\label{app:additional_qual}

\begin{figure*}[t]
    \centering
    \includegraphics[width=0.98\textwidth]{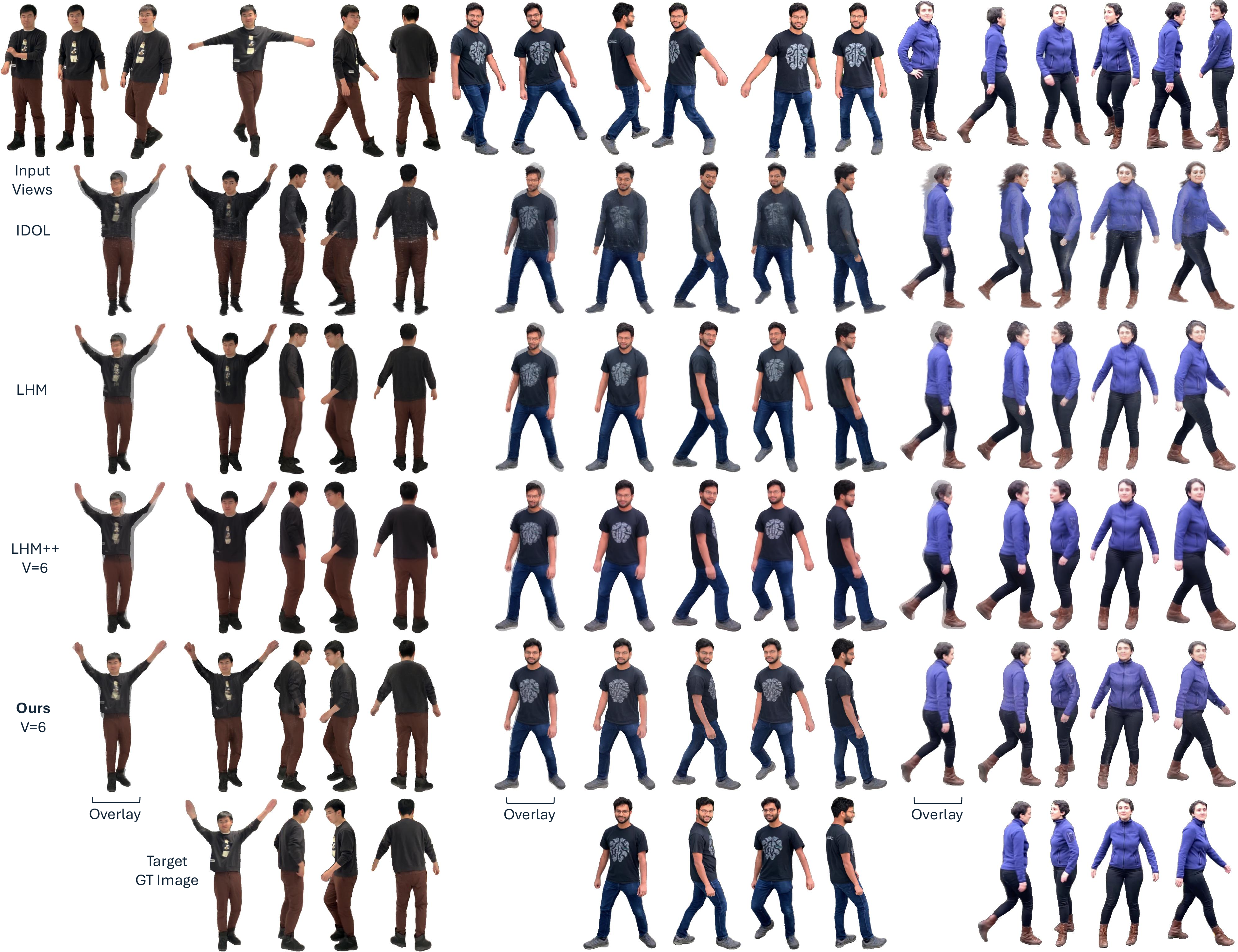}
    \caption{Additional qualitative comparisons of ARG-Avatar with the baselines on NeuMan. IDOL and LHM use the first view as input; LHM++ and ARG-Avatar use all six views. The first column of each group shows an overlay of the rendered and target images.}
    \label{fig:appendix_qual_neuman}
\end{figure*}

\begin{figure*}[t]
    \centering
    \includegraphics[width=0.98\textwidth]{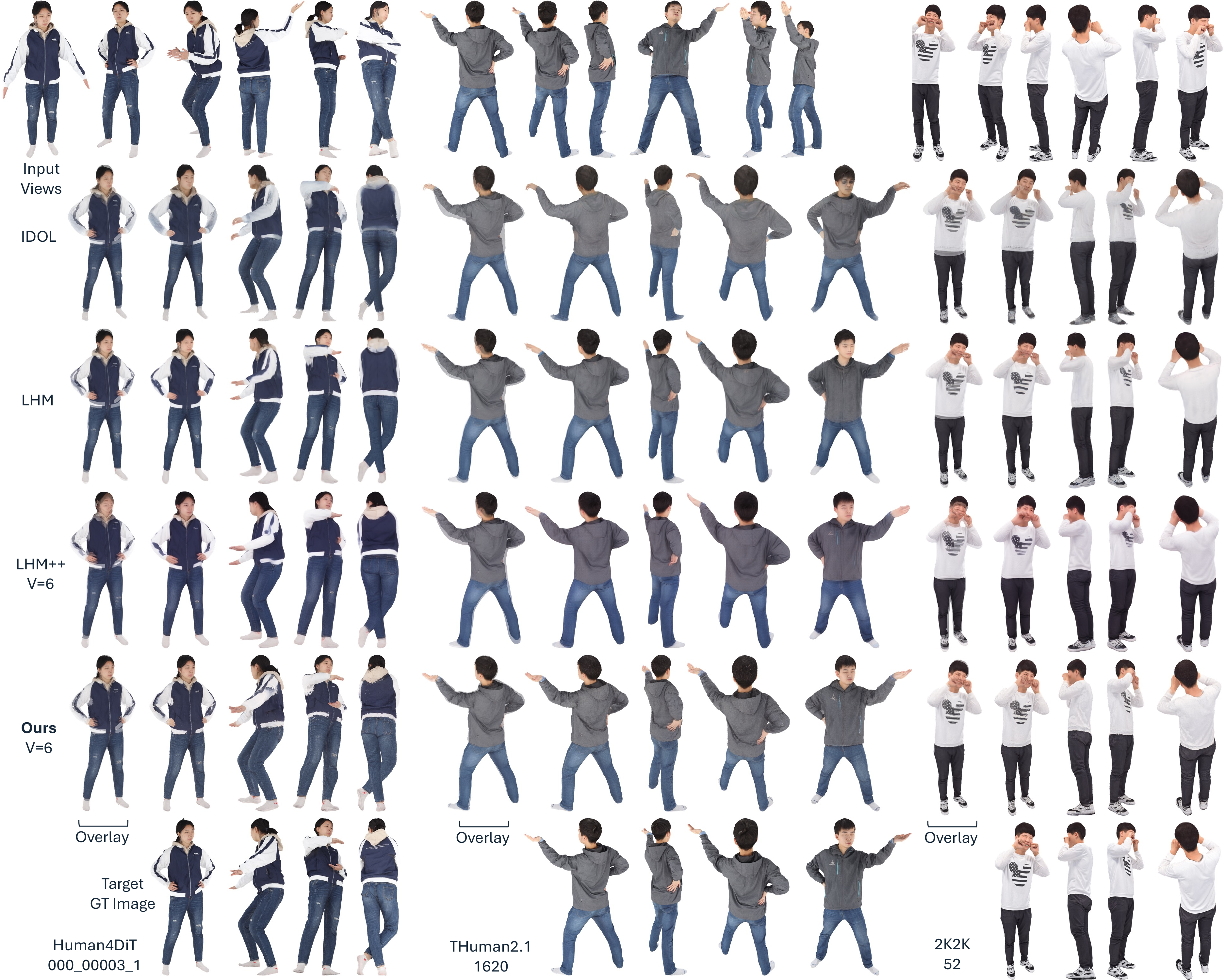}
    \caption{Additional qualitative comparisons of ARG-Avatar with the baselines on held-out validation subjects from Human4DiT, THuman2.1, and 2K2K. All samples are unseen during ARG-Avatar training, but 2K2K is \textbf{used to train} LHM and LHM++. Subject names are shown below their respective dataset names.}
    \label{fig:appendix_qual_1}
\end{figure*}

\begin{figure*}[t]
    \centering
    \includegraphics[width=0.98\textwidth]{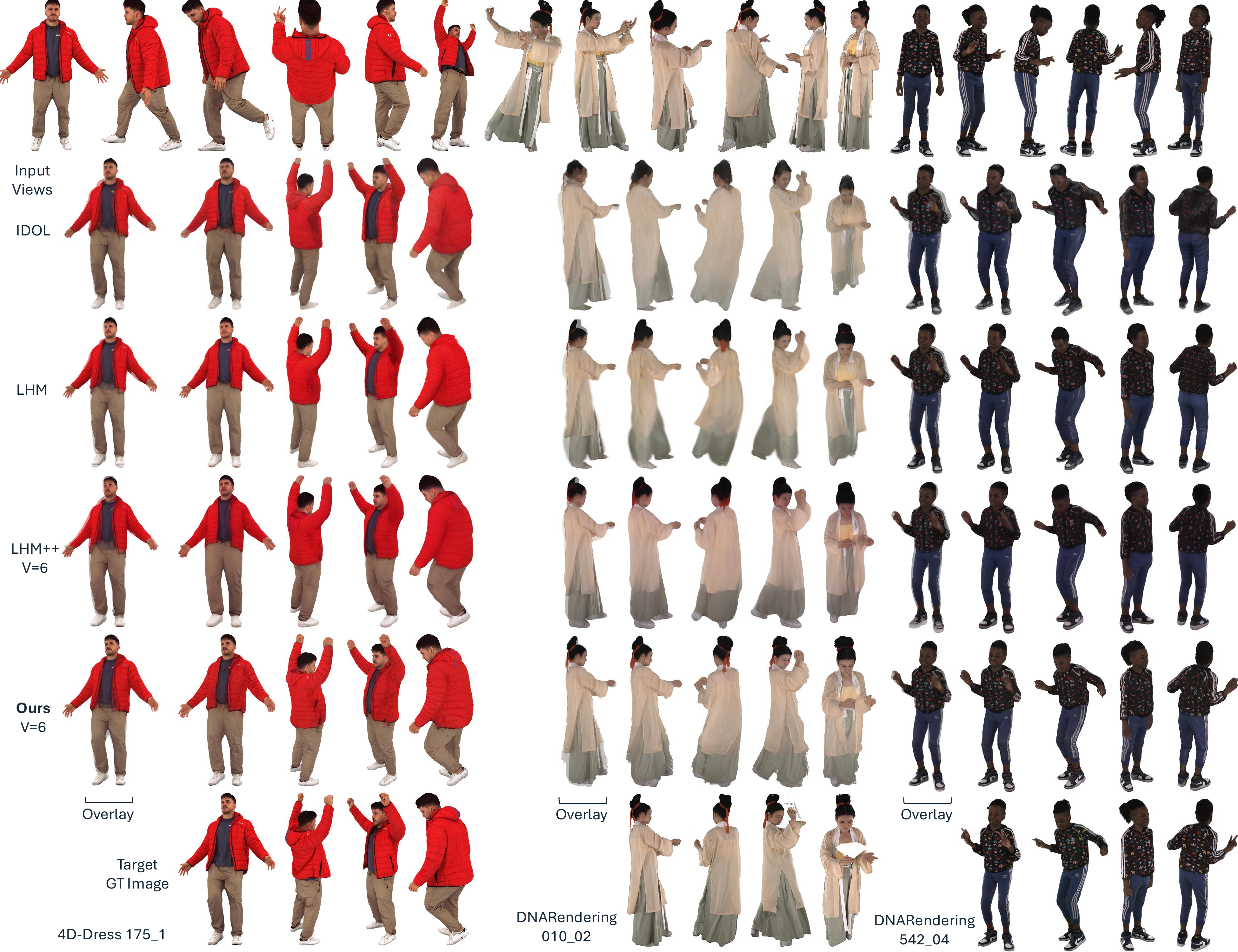}
    \caption{Additional qualitative comparisons of ARG-Avatar with the baselines on held-out validation subjects from 4D-Dress and DNARendering. Subject names are shown below their respective dataset names.}
    \label{fig:appendix_qual_2}
\end{figure*}

\cref{fig:appendix_qual_1,fig:appendix_qual_2} show additional qualitative comparisons on held-out validation subjects, complementing~\cref{fig:qualitative} in the main paper. \cref{fig:appendix_qual_neuman} shows additional comparisons on NeuMan, the in-the-wild benchmark that is never used in training.

\begin{samepage}
\section{Ablated ARGBlock Variants}
\label{app:argblock_variants}

\cref{fig:argblock_variants} shows the two ablated block variants evaluated in the ablation study, using compact ARGBlock diagrams that leave out the dropped modules.

\end{samepage}

\end{document}